\PassOptionsToPackage{numbers, sort&compress}{natbib}
\documentclass{article}

\usepackage[final]{corl_2026} %

\usepackage{amsmath}
\usepackage{amssymb}
\usepackage{bm}
\usepackage{import}
\usepackage{nicefrac}
\usepackage{graphicx}
\usepackage{hyperref}
\usepackage{tikz}
\usepackage{tabularx}
\usepackage{booktabs}
\usepackage{multirow}
\usepackage{threeparttable}
\usepackage{wrapfig}
\usepackage{transparent}
\usepackage[format=plain,labelformat=simple,labelsep=period,font=small,compatibility=false]{caption}
\usepackage[font=footnotesize,skip=3pt,subrefformat=parens]{subcaption}
\usepackage[basic,italic,defaultimath,nohbar,defaultmathsizes]{mathastext}

\usepackage[capitalize]{cleveref}
\crefname{section}{Sec.}{Secs.}
\Crefname{section}{Section}{Sections}
\Crefname{table}{Table}{Tables}
\crefname{table}{Tab.}{Tabs.}

\usepackage{xspace}
\usepackage{siunitx}
\definecolor{caveblue}{RGB}{0,68,136}
\definecolor{cavered}{RGB}{192,0,0}
\definecolor{plotred}{RGB}{255,64,64}

\definecolor{myred}{RGB}{255,64,64}
\definecolor{myblue}{RGB}{79,113,190}
\definecolor{myroyalblue}{RGB}{49,113,246}
\definecolor{mygreen}{RGB}{65,146,76}
\definecolor{mygold}{RGB}{242,166,62}
\definecolor{mydarkgray}{RGB}{147,147,147}
\definecolor{mylightgray}{RGB}{218,218,218}
\usetikzlibrary{shadows, shapes, positioning, arrows.meta, ext.paths.ortho, matrix, decorations.pathreplacing}

\makeatletter
\DeclareRobustCommand\onedot{\futurelet\@let@token\@onedot}
\def\@onedot{\ifx\@let@token.\else.\null\fi\xspace}
\makeatother

\def\eg{\emph{e.g}\onedot} 
\def\ie{\emph{i.e}\onedot} 
 
\def\etc{\emph{etc}\onedot}

\def\apriori{\emph{a priori}}

\usepackage{titlesec}
\titlespacing*{\section}{0pt}{4pt}{2pt}
\titlespacing*{\subsection}{0pt}{3pt}{1.5pt}
\titlespacing*{\subsubsection}{0pt}{2pt}{1pt}
\titlespacing*{\paragraph}{0pt}{0pt}{0.33em}
\titlespacing*{\subparagraph}{0pt}{0pt}{0.33em}

\def\papertitle{How Firm Should a Grasp Be?}
\title{\papertitle}

\def\maketitlesupplementary
   {
   \newpage
       {
        \centering
        \Large
        \textbf{\papertitle}\\
        \vspace{0.5em}Supplementary Material \\
        \vspace{1.5em}
       }
   }

\author{
  Matthew Beveridge\\
  Columbia University\\
  \texttt{beveridge@cs.columbia.edu}\\
  \And%
  Shree K.\ Nayar\\
  Columbia University\\
  \texttt{nayar@cs.columbia.edu}\\
}

\begin{document}
\maketitle
\vspace{-0.05in}
\vspace{-0.3in}
\begin{abstract}
    An ideal robot grasp is firm enough to securely handle an object, yet gentle enough to avoid damaging it. Achieving this balance requires knowledge of the object's material properties, such as its mass, elasticity, and surface friction. These properties, however, are seldom precisely known \textit{a priori}. In this work, we propose a visuotactile approach to estimating material properties in real time, during the process of grasping. Our method uses these estimated properties to determine the minimum grasp force required to handle the object. We contribute a new dataset of real-world objects (fruits and vegetables) with measured physical properties (shape, mass, elasticity, and friction), which we use to construct our force estimation model via simulations. We experimentally validate our approach to grasp force control using a robot with a parallel-jaw gripper. We demonstrate our system's ability to gently grasp a wide variety of objects, in each case adapting to their unique physical properties.
    \vspace{-0.05in}
\end{abstract}

{
  \footnotesize
}

\section{Gentle Grasping Requires Material Perception}
\label{sec:introduction}
Given an object and a desired interaction with it, the ideal grasp should exert no more force on the object than necessary to complete the task. While the notion of damage is often considered in catastrophic terms (\eg, a glass shattering), it can also be more subtle. For example, with excessive force, a fruit may bruise, a cardboard box might dent, or a plastic toy could crack or get scratched. In addition, there is a temporal element to damage: repeated interactions without care for an object's fragility can cause cumulative wear over time. In this work, we are interested in the \textit{gentle grasping} of objects, where the goal is to minimize the potential for damage during a grasp.

Since the material properties of an object are seldom precisely known \textit{a priori}, how do we, humans, determine how firm its grasp should be? Dynamic grasp adaptation is intuitive to us~\citep{paulunCenterSideBiases2014,paulunEffectsMaterialProperties2016} -- we use our vision to obtain priors on the object's material composition~\citep{flemingPerceptualQualitiesMaterial2013,flemingVisualPerceptionMaterials2014,flemingMaterialPerception2017}, and rely on tactile feedback to refine these priors while grasping the object~\citep{zoellerIntegrationPriorKnowledge2019,maielloHumansCanVisually2021}. Since some of the material cues we need to understand an object's physics cannot be known from vision alone, we estimate them via touch during the nascent stage of a grasp. We use these estimates to continuously adjust the forces we apply to the object and are thus able to gracefully handle it without conscious effort.

For a robot to gently grasp an object, it must either know the object's material properties beforehand or be able to infer them during the grasp. Unless the object is precisely engineered or measured in detail, the best the robot has to work with is a rough approximation of its properties. One method to refine this approximation is for the robot to interact with the object before attempting a grasp. That is, the robot ``probes'' the object with a sequence of exploratory actions to estimate, for instance, its hardness~\cite{liLearningGentleGrasping2025, nakaharaLearningGentleGrasping2025}. Such an approach is time-consuming and hence not suitable for applications where speed is critical. Instead, we propose to leverage the dynamic, ephemeral tactile signals that are sensed \textit{during a grasp} to estimate the object's properties in real time. The robot is therefore able to adjust the force it applies to the object on-the-fly, \textit{in situ}.

\begin{figure}
    \centering
    \hfill
    \begin{subfigure}[t]{0.22\linewidth}
        \centering
        \def\svgwidth{\linewidth}
        {
            \footnotesize
\begingroup%
  \makeatletter%
  \providecommand\color[2][]{%
    \errmessage{(Inkscape) Color is used for the text in Inkscape, but the package 'color.sty' is not loaded}%
    \renewcommand\color[2][]{}%
  }%
  \providecommand\transparent[1]{%
    \errmessage{(Inkscape) Transparency is used (non-zero) for the text in Inkscape, but the package 'transparent.sty' is not loaded}%
    \renewcommand\transparent[1]{}%
  }%
  \providecommand\rotatebox[2]{#2}%
  \newcommand*\fsize{\dimexpr\f@size pt\relax}%
  \newcommand*\lineheight[1]{\fontsize{\fsize}{#1\fsize}\selectfont}%
  \ifx\svgwidth\undefined%
    \setlength{\unitlength}{97.19999313bp}%
    \ifx\svgscale\undefined%
      \relax%
    \else%
      \setlength{\unitlength}{\unitlength * \real{\svgscale}}%
    \fi%
  \else%
    \setlength{\unitlength}{\svgwidth}%
  \fi%
  \global\let\svgwidth\undefined%
  \global\let\svgscale\undefined%
  \makeatother%
  \begin{picture}(1,0.74074079)%
    \lineheight{1}%
    \setlength\tabcolsep{0pt}%
    \put(0.37037038,0.62962974){\color[rgb]{0,0,0}\makebox(0,0)[t]{\lineheight{1.25}\smash{\begin{tabular}[t]{c}``Pepper''\end{tabular}}}}%
    \put(0,0){\includegraphics[width=\unitlength,page=1]{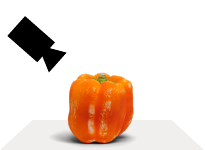}}%
  \end{picture}%
\endgroup%

        }
        \vspace{-0.155in}
        \subcaption*{\textit{Observe}}\label{fig:phases:observe}
    \end{subfigure}
    \hfill
    \begin{subfigure}[t]{0.22\linewidth}
        \centering
        \def\svgwidth{\linewidth}
        {
            \footnotesize
            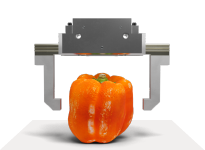
        }
        \vspace{-0.155in}
        \subcaption*{\textit{Approach}}\label{fig:phases:approach}
    \end{subfigure}
    \hfill
    \begin{subfigure}[t]{0.22\linewidth}
        \centering
        \def\svgwidth{\linewidth}
        {
            \footnotesize
            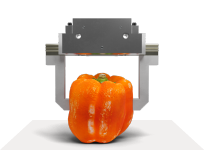
        }
        \vspace{-0.155in}
        \subcaption*{\textit{Grip}}\label{fig:phases:grip}
    \end{subfigure}
    \hfill
    \begin{subfigure}[t]{0.22\linewidth}
        \centering
        \def\svgwidth{\linewidth}
        {
            \footnotesize
            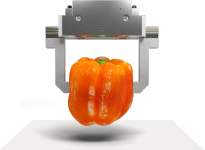
        }
        \vspace{-0.155in}
        \subcaption*{\textit{Lift}}\label{fig:phases:lift}
    \end{subfigure}
    \hfill
    \vspace{-0.08in}
    \caption{%
        \textbf
        {Phases of a Grasp.} We consider a grasp to comprise four stages: observe, approach, grip, and lift. Our work exploits the short-lived (fraction of a second) haptic signals during the grip and lift phases to refine class-level material properties (from observation) to obtain instance-level estimates. During this refinement, we continuously adjust the applied force to ensure it is just above the minimum force needed to handle the object.
    }
    \label{fig:phases}
    \vspace{-0.2in}
\end{figure}

In a typical robotic grasping task, there are four initial phases as illustrated in \cref{fig:phases}. The first is \textit{observe}, where, from a distance, the robot can visually identify the class of the object, its shape, and its location. During the \textit{approach} phase, the robot positions itself at a suitable grasp location determined by the object's geometry and pose. Next, we have \textit{grip}, where the gripper squeezes the object to take hold. Finally, the robot \textit{lifts} the object before executing a goal task, such as moving it to a new location or mating it with another object. In our work, we focus on the \textit{observe}, \textit{grip}, and \textit{lift} phases. The \textit{observe} phase reveals the class of the object (``pepper'' in  \cref{fig:phases}), giving us a prior on its properties (mass, elasticity, friction, \etc). During the \textit{grip} and \textit{lift} phases, we use haptic feedback to continuously refine these priors and control the force applied to the object. For \textit{approach}, we assume a reasonable grasp pose is provided by an existing planner~\cite{xuMinimalWorkGrasp2020,huangDefGraspNetsGraspPlanning2023}.

While haptic signals during a grasp encode an object's material properties, different combinations of material parameters could lead to the same haptic signals. Therefore, material estimation can be viewed as a challenging inverse problem. To develop our solution, we have chosen the domain of fruits and vegetables. This is an interesting domain as two samples, even when of the same class, can be expected to differ in their properties (\eg, a raw and a ripe avocado). We first collected a dataset, called \textit{Squash}, that includes material measurements and 3D scans of 270 samples across 52 different classes of fruits and vegetables.\footnote{The dataset is available through the \href{https://cave.cs.columbia.edu/projects/categories/project?cid=Physics-Based+Vision&pid=How+Firm+Should+a+Grasp+Be}{project webpage}.} From \textit{Squash}, we derived class-level property distributions, which reduce the search space for instance-level properties.

Using \textit{Squash}, we simulated numerous object grasps, with augmentations of material properties to span real-world variability. We trained a sequence model on the evolution of the simulated haptic data (\ie, tactile images and load cell measurements) to distill instance-level material properties from class distributions. We have integrated our visuotactile model for force control into a robot's operating system, where it runs at 250~\unit{\Hz}. We used a 6 DoF arm and a parallel-jaw gripper with GelSight tactile sensors to validate our approach. Our system achieved more than a 2$\times$ reduction in excess applied force for a successful grasp compared to non-adaptive baselines.

\section{Related Work}
\label{sec:related_work}

\paragraph{Class-Level Material Recognition from a Distance.} 
For grasping, it is beneficial to recognize an object's material class (\eg, peach) passively during the \textit{observation} stage. However, precise class-level visual recognition (\ie, peaches vs.\ nectarines) requires elaborate optical measurements, such as the reflectance distribution function (BRDF)~\citep{zhangReflectanceHashingMaterial2015,danaReflectanceTextureRealworld1999,weinmannMaterialClassificationBased2014}, which is hard in dynamic, cluttered robotics settings. Hence, class recognition is most often done from appearance alone~\citep{adelsonSeeingStuffPerception2001,sharanRecognizingMaterialsUsing2013,bellMaterialRecognitionWild2015,danaComputationalTexturePatterns2018,liuBoWCNNTwo2019}, using filter banks, statistical analysis, and deep learning. These approaches do not directly relate appearance to material properties, which has been the goal of more recent work~\citep{schwartzRecognizingMaterialProperties2020,purriTeachingCamerasFeel2020,beveridgeHierarchicalMaterialRecognition2025}. In our work, we rely on appearance to identify class-level material properties during \textit{observation}, while recognizing that additional sensing modalities are needed to resolve instance-level variation within a class.

\paragraph{Instance-Level Material Recognition through Excitation.}
During the \textit{grip} and \textit{lift} stages, haptic feedback provides material-related signals for the specific object of interest (\eg, \textit{this} peach has \textit{this} hardness). In the general context of measuring physical properties, exciting objects with ultrasound and physical impact have been shown to reveal useful material cues~\cite{klatzkyPerceptionMaterialContact2000,fangFingertipSensorAlgorithms2023,zhangAnalyzingPhysicalImpacts2023}. In the realm of robotics, vision-based tactile sensors have been used to provide detailed insights related to contact mechanics. For example, GelSight~\cite{johnsonRetrographicSensingMeasurement2009,johnsonMicrogeometryCaptureUsing2011,yuanGelSightHighResolutionRobot2017} and DIGIT~\cite{lambetaDIGITNovelDesign2020} have been used for hardness estimation~\cite{yuanShapeindependentHardnessEstimation2017} and shear/slip detection~\cite{yuanMeasurementShearSlip2015}. Beyond optical sensors, piezoresistive~\cite{bhirangiReSkinVersatileReplaceable2022,huangFlexiTacLowCostOpenSource2026} and magnetic tactile skins~\cite{pattabiramanEFleshHighlyCustomizable2025}, which measure pressure distributions, have been integrated into grippers for fine-grained visuotactile manipulation~\cite{huang3DViTacLearningFineGrained2024,zhuTouchWildLearning2025}. Cross-modal methods have also been developed to relate visual appearance to tactile properties~\cite{yuanConnectingLookFeel2017,liConnectingTouchVision2019,purriTeachingCamerasFeel2020,yuDefGraspRobotGrasping2023,ruanVisionTactileFusionBased2023}. In our work, we use GelSight to image contact geometry, shear, and slip during a grasp. However, our approach can apply to various tactile sensors.

\paragraph{Approaches to Gentle Grasping.}
A classical approach to avoiding object damage is to regulate contact force via impedance control~\cite{hoganImpedanceControlApproach1985}, or to adjust grasp force reactively using tactile feedback~\cite{liRealTimeAutonomousGrasping2018,dharbaneshwerContactAreaBasedModeling2021,davellaAutonomousSoftGrasping2022}, \eg, upon detecting incipient slip~\cite{jamesSlipDetectionGrasp2021,juddSlipAnticipationGrasping2022}. For this purpose, tactile sensors have also been integrated into soft robotic hands due to their passive compliance~\cite{leSafeGraspingForce2020,dongGSGGranarySoft2021}. Purely vision-based methods~\cite{leSynthesizingGrasps3D2021,leDeformationAwareDataDrivenGrasp2022,linPlanningSpatialTemporalAbstraction2023,zhangAdaptiGraphMaterialAdaptiveGraphBased2024,zhangParticleGridNeuralDynamics2025,jiangPhysTwinPhysicsInformedReconstruction2025,zhangBoxTwinLearningElastoplastic} have also demonstrated successful deformation-aware grasping, but are shown to be more effective when fused with tactile sensing~\cite{calandraMoreFeelingLearning2018,cuiGraspStateAssessment2020}. More recently, human and robot demonstrations have enabled end-to-end gentle grasping of objects~\cite{luoAdaptiveTactileInteraction2024,liLearningGentleGrasping2025, nakaharaLearningGentleGrasping2025,zhuTouchWildLearning2025,wiTactAlignHumantoRobotPolicy2026}, and diffusion-based policies conditioned on tactile signals have been used for force-aware manipulation of fragile objects~\cite{helmutTactileConditionedDiffusionPolicy2025}. Nevertheless, these methods are not real-time, do not expose inferences about the object's material, and only yield predictions of an ``acceptable'' grasp force based on heuristics. 

\paragraph{3D Object Datasets.}
Several large-scale 3D object datasets have been developed for robotic manipulation~\cite{calliBenchmarkingManipulationResearch2015,liangPointNetGPDDetectingGrasp2019,mousavian6DOFGraspNetVariational2019,yanLearning6DOFGrasping2018,fangGraspNet1BillionLargeScaleBenchmark2020} and broader 3D vision tasks~\cite{changShapeNetInformationRich3D2015,downsGoogleScannedObjects2022,deitkeObjaverseUniverseAnnotated2023}. These datasets provide geometry and appearance but do not include measured material properties. The recent work by \citet{caoSOPHYLearningGenerate2025} uses annotated simulated objects with material parameters, where the parameter values are assigned using material-category labels rather than measurements made on individual instances. For deformable objects, smaller datasets exist that include detailed annotations of the deformation process through finite element analysis~\cite{leSynthesizingGrasps3D2021,huangDefGraspSimPhysicsBasedSimulation2022}. One of our contributions is \emph{Squash}, a dataset that pairs 3D scans of fruits and vegetables with per-instance measurements of material properties (elasticity, mass, and surface friction parameters). It is the first 3D object dataset specifically designed for deformable object simulation with measured properties. Beyond its utility in robotics, we expect \emph{Squash} to be a valuable resource for simulation and rendering in computer graphics and animation.
\section{The Physics of Grasping a Material}
\label{sec:physics}
\begin{wrapfigure}{r}{2.4in}
    \vspace{-0.475in}
    \centering
    \def\svgwidth{\linewidth}
    {
        \scriptsize
        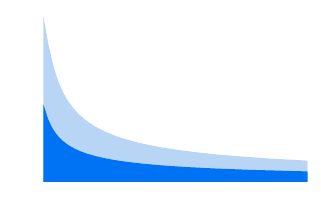
    }
    \vspace{-0.15in}
    \caption{
        \textbf{Soft Objects Require Less Grasp Force.}
        For a soft object, the contact area $A$ grows more rapidly with applied force $F$ than it does for a hard object. As $A$ grows, an additional load, $\tau A$, where $\tau$ is the object's material shear strength, supplements $F$ to resist the effect of gravity (\cref{eq:soft-force-equilibrium}). The theoretical reduction in minimum-required grasp force due to this additional load is shown here as a function of the elastic modulus. The object and its shear strength $\tau$ are kept constant in the plot, but the ratio of mass $m$ to friction $\mu$ is varied. The greatest benefit (\ie, the possible reduction in grasp force) is for light, pliable objects.
    }\label{fig:soft-vs-rigid}
\end{wrapfigure}

If we know the material properties of an object, then what minimal grasp force should we use to lift it? Na\"ively, the total grasp force $F$ required to suspend a rigid body of mass $m$ at equilibrium is $F = mg$, \ie, Newton's second law. If we assume that the force due to friction is independent of the mass and the contact area, we would get the familiar modification: $\mu F = mg$, where $\mu$ is the coefficient of friction. Coulomb observed that the effect of friction is primarily due to the interlocking of surface asperities, which he assumed to be infinitesimally small. This, however, is an approximation. \citet{bowdenFrictionLubricationSolids2001} noted that friction capacity can vary under pressure, thereby acting as an additional load, especially when the interacting bodies are soft. The effect of this additional load is visualized in \cref{fig:soft-vs-rigid}, where we see that soft objects require a lower grasp force than rigid ones, since they deform and create a larger contact area under the same load. Formally, this yields a new equilibrium:
\begin{align}
    \mu F + \tau A_{\text{real}} = mg, \label{eq:soft-force-equilibrium}
\end{align}
where $\tau$ is the interfacial shear strength, and $A_{\text{real}}$ is the contact area over which surface asperities interact. For rigid bodies, $A_\text{real}$ is considerably smaller than the apparent contact region $A_\text{app}$ (on the macroscale), as only the peaks of the asperities interlock~\cite{hanaorStaticFrictionFractal2016}.

If the object or the gripper is soft, we can say $A_{\text{real}} \approx A_{\text{app}}$, since the elastic deformation fills the cavities between the summits of surface asperities. In our context, $A_\text{app}$ can be directly observed through tactile sensing. Even so, $F$ and $A_\text{app}$ are directly coupled, and without a mapping between them, we must \textit{search} for the force that results in the equilibrium given by \cref{eq:soft-force-equilibrium}. To \textit{predict} this force of equilibrium, we must also predict how the apparent contact area will change with respect to the load force. Finite element analysis~(FEA) can accurately estimate this contact area, assuming the material properties of the object and gripper are known, \apriori. In our case, property estimates change during the course of a grasp, necessitating an expensive FEA reevaluation on each update, which would limit the rate at which we can adapt grasp force.

Instead, by using a Hertzian contact approximation~\cite{hertz1896contactElasticSolids,hertz1896contactHardness}, $A_\text{app}$ can be modeled using a closed-form expression. If we approximate the fingertips of a robotic gripper as near-planar, and an object as locally spherical, then applying a normal force $F$ yields:
\begin{align}
    A_{\text{app}}(F) & = \pi \left ( \frac{3FR^*}{4E^*}  \right )^{\nicefrac{2}{3}} \hspace{-0.15in}, \,\,
    \text{given} \;\; \frac{1}{E^*} = \frac{1-\nu_{\text{obj}}^2}{E_{\text{obj}}} + \frac{1-\nu_{\text{grip}}^2}{E_{\text{grip}}} \;\;
    \text{and} \;\; R^* = \frac{R_{\text{obj}} \, R_{\text{grip}}}{R_{\text{obj}} + R_{\text{grip}}},
    \label{eq:hertz-contact-area}
\end{align}
where $E^*$ is the effective elastic modulus between the object and the gripper, $R^*$ is the effective radius of curvature between the object and the gripper at the contact interface, and $\nu$ is the Poisson ratio. Given the mass $m$, elasticity $E$, surface friction $\mu$, and interfacial shear strength $\tau$ of an object, we can find $A_{\text{app}}$ using \cref{eq:hertz-contact-area} and substitute in \cref{eq:soft-force-equilibrium} to estimate the minimum force needed to lift the object. But, for an arbitrary object, we at best know a range of plausible values for $(E, m, \mu, \tau)$ based on its class. Our goal, therefore, is to develop a learning-based approach to refine $(E, m, \mu, \tau)$ through temporal changes in haptic feedback, which we then use to servo-control the applied force.

\begin{figure}
    \centering
    \begin{subfigure}[b]{0.27\linewidth}
        \centering
        \includegraphics[width=\linewidth]{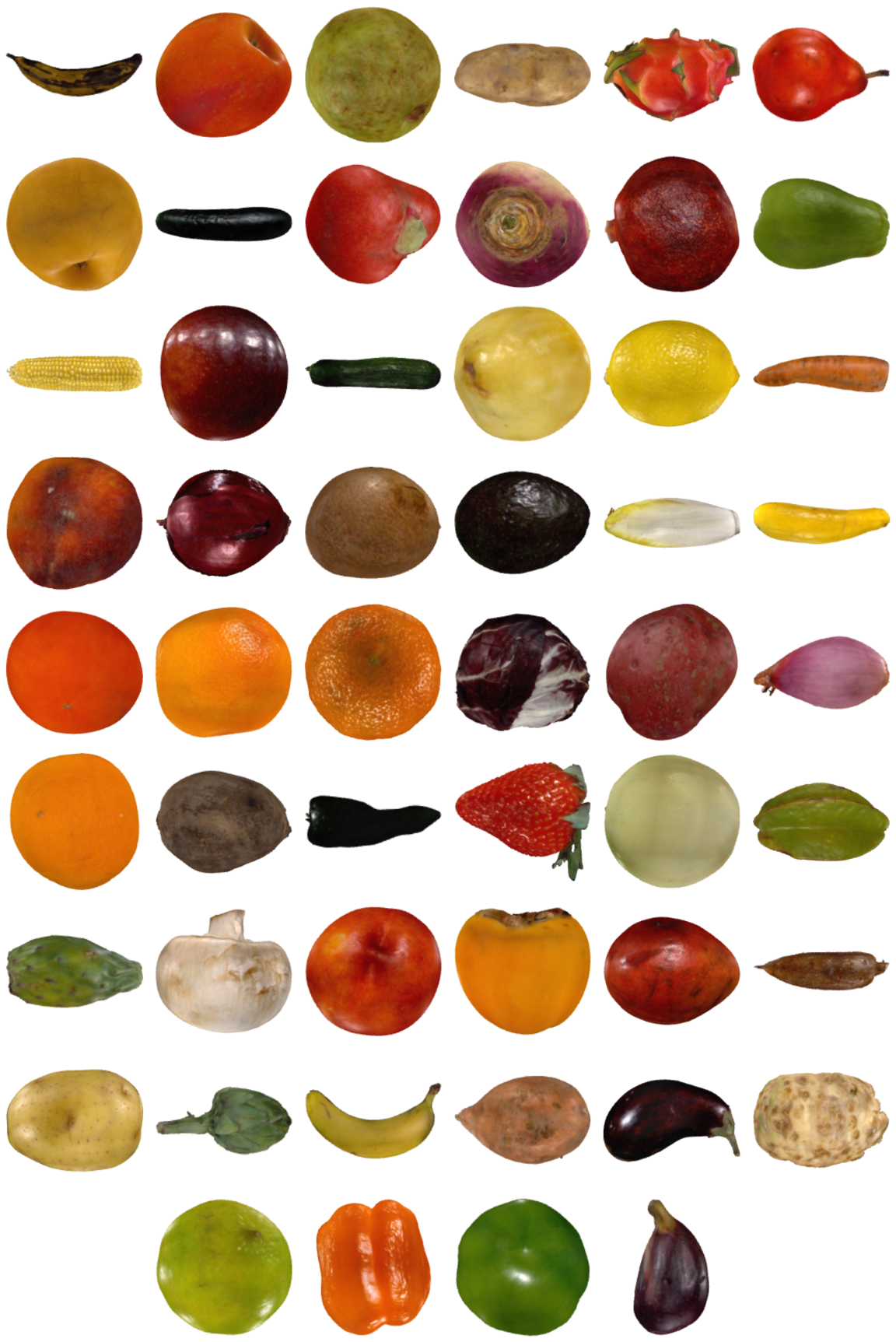}
        \subcaption{52 Material Classes}\label{fig:squash-dataset:classes}
        \vspace{-0.05in}
    \end{subfigure}
    \hfill
    \begin{subfigure}[b]{0.26\linewidth}
        \centering
        \def\svgwidth{\linewidth}
        {
            \scriptsize
            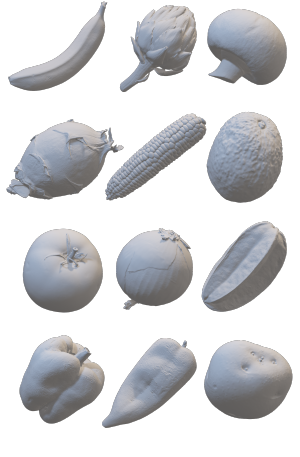
        }
        \vspace{-0.15in}
        \subcaption{Example 3D Scans}\label{fig:squash-dataset:scans}
        \vspace{-0.05in}
    \end{subfigure}
    \hfill
    \hspace{-0.125in}
    \begin{subfigure}[b]{0.45\linewidth}
        \centering
        \def\svgwidth{\linewidth}
        {
            \scriptsize
            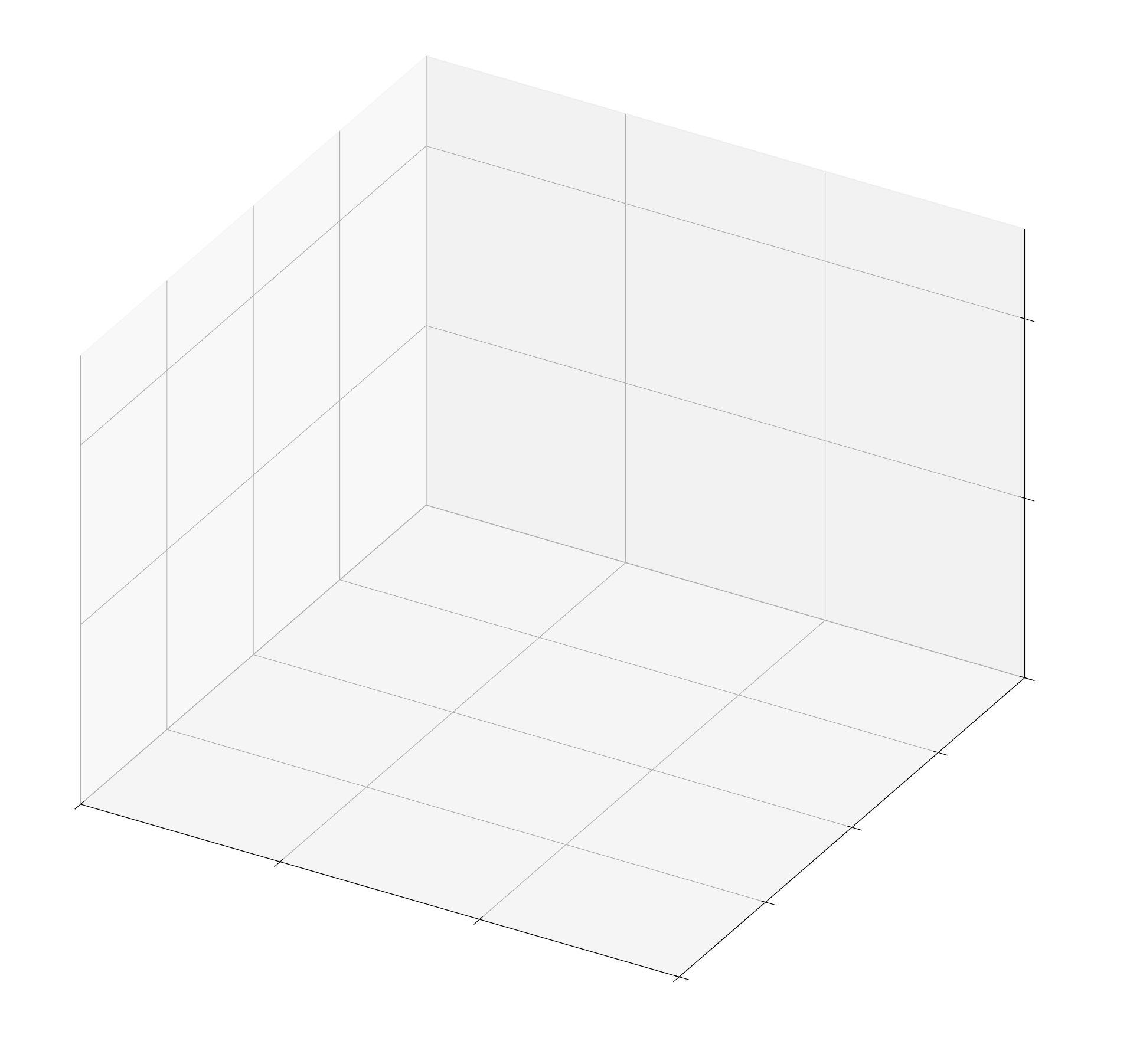
        }
        \vspace{-0.175in}
        \subcaption{Mechanical Property Distributions}\label{fig:squash-dataset:mech-props}
        \vspace{-0.05in}
    \end{subfigure}
    \caption{
        \textbf{The \textit{Squash} Dataset.} A dataset of real-world 3D object meshes and material property measurements. (a)~\textit{Squash} is composed of 270 instances across 52 classes of produce (fruits and vegetables). (b)~We scanned each object using structured light to measure its geometry and texture at high resolution. (c)~We additionally measured the material properties of each object (elastic modulus, mass, volume, and surface friction), where each ellipsoid represents the distribution of the properties within a class.
        \textit{Squash} is a first-of-its-kind dataset that captures the natural variability of object properties, which can be used for physically-accurate simulation.
    }\label{fig:squash-dataset}
    \vspace{-0.2in}
\end{figure}

\section{\textit{Squash}: A Dataset of 3D Object Assets and their Physical Properties}
\label{sec:dataset}
To learn our visuotactile model for gentle grasping, we require a dataset of deformable objects with known physical properties. Deformable objects are interesting in two respects. First, they are more prone to damage due to a grasp that is too firm (\eg, a peach will bruise). Second, as outlined in \cref{sec:physics}, the contact area between the gripper and the object changes as they both deform, thereby requiring continuous control of the minimum grasp force. We have developed \textit{Squash} (\cref{fig:squash-dataset}), a comprehensive dataset of 3D scans and physical property measurements of various fruits and vegetables. \textit{Squash} contains 270 instances of produce spread across 52 classes (capturing intraclass variability). All 52 classes are shown in \cref{fig:squash-dataset:classes}, and a few examples of the 3D scans are shown in \cref{fig:squash-dataset:scans}. Using the offline probing methods described below, we measured the elastic modulus, mass, volume, and surface friction parameters of each instance. A visualization of the measured properties is shown in \cref{fig:squash-dataset:mech-props}, where each ellipsoid represents the distribution of the properties within a class.

\paragraph{Characterizing the Properties of an Object.}
For each object, we first measured its mass and Shore hardness using a durometer~\cite{gentRelationIndentationHardness1958,mixStandardizedPolymerDurometry2011}. From its 3D scan, we then computed its volume. Next, using a robot arm and a parallel-jaw gripper affixed with load cells and tactile fingertips (detailed in the supplemental material), we measured the object's elasticity and friction capacity by performing a compression test and a tribological test, respectively. In the compression test, with the object at rest, we recorded the applied force $F_\text{load}$, contact area $A$, and indentation into the object $L$, while closing the gripper jaws on the object. We modeled the local surface of the object as quasi-spherical and the gripper jaws as parallel plates. Then, for any deformation of the object, the effective modulus $E^*$ follows from \cref{eq:hertz-contact-area}: $\smash{E^*= \frac{3 F}{4 L} \sqrt{\frac{\pi}{A}}}$, since from Hertzian mechanics, the effective curvature of the contact interface is $\smash{R^* = \frac{A}{\pi L}}$. We have many samples of effective modulus $E^*$ (during closure), and we averaged them for our final measurement of $E^*$. We then recovered the object's elastic modulus $E$ from $E^*$ by separately measuring the elastic modulus of the tactile sensor. We report $E$ as a function of the Poisson ratio $\nu$, since we cannot directly observe the equatorial deformation of the object.

For the tribological test, we began with the object lifted, then increased the gripper opening until slip was detected by the tactile sensor. Each trial yielded one $(F_\text{slip}, A_\text{slip})$ pair, where $F_\text{slip}$ is the applied force just before slip, and $A_\text{slip}$ is the contact area at that onset. These measurements are related through \cref{eq:soft-force-equilibrium}: $\mu F_{\text{slip}} + \tau A_{\text{slip}} = mg$. We collected several trials per object, each at a different location on its surface, and solved the over-determined linear system in $(\mu, \tau)$ by least squares. By sampling distinct locations on the object, we captured local deviations in surface curvature. This gave us different couplings between $F_\text{slip}$ and $A_\text{slip}$, which enabled us to decouple the effects of $\mu$ and $\tau$. Since \textit{Squash} includes a large sampling of shapes, masses, elasticity parameters, and friction parameters, we can use it to generate diverse grasp simulations for model training.
\begin{figure}
    \centering
    \def\svgwidth{\linewidth}
    {
        \footnotesize
        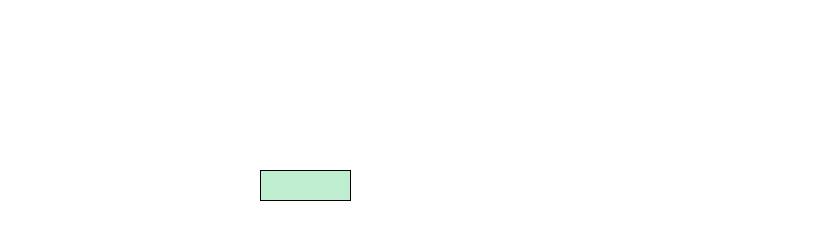
    }
    \vspace{-0.2in}
    \caption{
        \textbf{Framework for Gentle Grasping.}
        From \textit{observation}, we identify the object's class to get a class-conditioned material prior and an initial minimum-required grasp force estimate. Upon first contact, our visuotactile likelihood model $\psi$ (detailed in \cref{fig:model}) begins to refine these priors into instance-level posteriors from temporal tactile feedback. The minimum-required grasp force is continuously reestimated in the process, and applied in real time. The result is a gentle grasp that adapts to the object instance, \textit{in situ}.
    }\label{fig:framework}
    \vspace{-0.25in}
\end{figure}

\section{Material-Aware Robotic Grasping}
\label{sec:method}
\cref{fig:framework} illustrates our online, closed-loop grasping framework for estimating
material properties $\theta=(E, m, \mu, \tau)$ to determine the minimum necessary grasp force $F_{\min}$ for an object. The material prior distribution $p_c(\theta) \equiv p(\theta \mid c)$ for the object class $c$, identified during \textit{observation}, yields an initial estimate of the minimum grasp force $F_{\min,c}$ via \cref{eq:soft-force-equilibrium}. The \textit{visuotactile likelihood model}, $\psi$, then refines this class prior during contact with the object, giving an online posterior belief over material properties at each time $t$. As the robot begins to grip the object, tactile images $\smash{I^{(t)}}$ are used to update the elasticity estimate $\smash{E^{(t)}}$ from temporal changes in indentation $\smash{L^{(t)}}$ and contact area $\smash{A^{(t)}}$ relative to the applied load $\smash{F_{\text{load}}^{(t)}}$.\footnote{While our implementation uses GelSight tactile images, the model $\psi$ can utilize other tactile sensor outputs, \eg, binary contact maps and pressure maps. For our purposes, we refer to all these outputs as tactile images.} As the robot begins to lift the object, shear $\smash{V^{(t)}}$ and slip $\smash{U^{(t)}}$ in the tactile images are used to update the estimated mass $\smash{m^{(t)}}$ and friction parameters $\smash{\mu^{(t)}}$ and $\smash{\tau^{(t)}}$. The model $\psi$ therefore continuously refines the material estimate $\smash{\theta^{(t)}}=(E^{(t)},m^{(t)},\mu^{(t)},\tau^{(t)})$, and the controller accordingly adjusts the applied grasp force $\smash{F_{\text{load}}^{(t)}}$ to achieve the current minimum-force estimate $\smash{F_{\min}^{(t)}}$. As additional tactile observations arrive, the posterior concentrates around the instance-level material properties $\theta_i$, yielding the desired object-specific force estimate $F_{\min,i}$ for gentle grasping. 

\paragraph{Learning Material from Simulated Grasps.}
To develop our visuotactile model $\psi$, we first designed a simulator to generate high-fidelity tactile data for training. We built upon IPC~\cite{liIncrementalPotentialContact2020,huangGIPCFastStable2024,huangStiffGIPCAdvancingGPU2025} for
\begin{wrapfigure}{r}{2.4in}
    \vspace{0.05in}
    \centering
    \def\svgwidth{\linewidth}
    {
        \footnotesize
        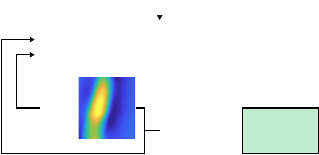
    }
    \vspace{-0.15in}
    \caption{
    \textbf{Visuotactile Model $\boldsymbol{\psi}$ (shaded green) for Material Estimation and Force Control.}
    Tactile images $\smash{I^{(t)}}$ are first encoded and passed with the applied load $\smash{F_{\text{load}}^{(t)}}$ to the continuous-time sequence model (CfC), yielding the hidden state $\smash{\bm{h}^{(t)}}$. The likelihood head uses this representation to score candidate material samples, $\smash{S_j^{(t)} = S_\psi(\bm{h}^{(t)}, \theta_j)}$, where samples $\theta_j$ are drawn from the class prior $p_c(\theta)$ for an object's material. Normalizing these scores yields the weighted belief $\{\smash{\theta_j},\smash{\alpha_j^{(t)}}\}_{j=1}^{N}$. The weighted samples map to force hypotheses whose posterior expected minimum force, $\smash{F_{\min}^{(t)}}$, sets the target force for the gripper controller in real time.
    }\label{fig:model}
    \vspace{-0.15in}
\end{wrapfigure}
collision physics, which has been shown to accurately model real-world grasps of deformable objects~\cite{kimIPCGraspSimReducingSim2Real2022,nguyenTacExGelSightTactile2024}, by modifying it to also model the adhesive effect of $\tau$ in \cref{eq:soft-force-equilibrium}. From the empirical probing tests in \cref{sec:dataset}, we observed a Pearson correlation of 0.94 between the true minimum grasp force and one modeled by our simulations. We then synthesized 200k training grasp sequences using a parallel-jaw gripper and the object meshes and material property distributions in \textit{Squash}. Each sequence contains temporal measurements of each jaw's tactile image $\smash{I^{(t)}}$ and applied normal load $\smash{F_{\text{load}}^{(t)}}$. To make real-world estimation robust, we replicated our real-world load cell and GelSight sensor characteristics in simulation, added noise to their measurements, and supplemented the \textit{Squash} material properties by widely sampling simulated materials. We then trained $\psi$ (\cref{fig:model}) to encode these temporal haptic observations. At each timestep $t$, the tactile image $\smash{I^{(t)}}$ is embedded by an image encoder, paired with $\smash{F_{\text{load}}^{(t)}}$, and passed to a closed-form continuous-time neural network (CfC)~\cite{hasaniClosedformContinuoustimeNeural2022} to produce a recurrent hidden state $\smash{\bm{h}^{(t)}}$. A CfC was chosen for its inference speed, ability to handle non-uniformly sampled robot data, and favorable scaling behavior when modeling complex dynamics. At inference, $\smash{\bm{h}^{(t)}}$ summarizes the observations accumulated through $t$ and is used by a learned likelihood head to evaluate candidate material parameters. We trained the image encoder, CfC, and likelihood head with a hybrid contrastive-regression loss over the simulated grasps.
Further details are provided in the supplemental material.

\paragraph{Gentle Grasping from Haptic Observations.}
The hidden state $\smash{\bm{h}^{(t)}}$ summarizes the observation history and parameterizes its cumulative likelihood under candidate material parameters $\theta$: $\smash{p_\psi (I^{(0:t)}, F_{\text{load}}^{(0:t)} \mid \theta)}$. The likelihood head thus gives the unnormalized cumulative log-likelihood,
\begin{align}
    S_\psi (\bm{h}^{(t)}, \theta) = \log p_\psi (I^{(0:t)}, F_{\text{load}}^{(0:t)} \mid \theta) + C^{(t)},
\end{align}
where $C^{(t)}$ is independent of $\theta$. Combining this with the class prior yields the posterior material belief,
$\smash{
    p (\theta \mid I^{(0:t)}, F_{\text{load}}^{(0:t)}) \propto p_c(\theta) \exp(S_\psi (\bm{h}^{(t)}, \theta)),
}$
which we approximate by importance sampling:
\begin{align}
    p(\theta \mid I^{(0:t)}, F_{\text{load}}^{(0:t)}) \approx \sum\nolimits_j \alpha_j^{(t)} \delta(\theta-\theta_j),
    \quad
    \alpha_j^{(t)} = \frac{\exp ( S_\psi(\bm{h}^{(t)}, \theta_j) )}{\sum_k \exp ( S_\psi( \bm{h}^{(t)}, \theta_k)  )} ,
\end{align}
where $\smash{\alpha_j^{(t)}}$ is the normalized importance weight of sample $\smash{\theta_j \sim p_c(\theta)}$. Samples are drawn once per grasp from the class prior $p_c(\theta)$ and held fixed; as haptic observations accumulate, only their weights are updated.
Each sample defines a candidate contact model and grasp feasibility condition~\cite{ferrariPlanningOptimalGrasps1992,millerGraspitVersatileSimulator2004}. While lifting the object, feasibility reduces to whether the frictional support exceeds the object's weight. The minimum required grasp force for a candidate material vector is therefore:
\begin{align}
    f(\theta) = \min\nolimits_{F \ge 0} \left\{ F : \mu F + \tau A(F;E) \ge mg \right\},
    \label{eq:force-hypothesis}
\end{align}
given by \cref{eq:soft-force-equilibrium}, where $A(F;E)$ follows from \cref{eq:hertz-contact-area}. Mapping each sample through $f$ gives the force hypothesis $\smash{F_{\min,j}}=\smash{f(\theta_j)}$, and the weighted set $\{\smash{F_{\min,j}},\smash{\alpha_j^{(t)}}\}$ represents the current uncertainty in the required grasp force. The controller then tracks the posterior expected minimum force $\smash{F_{\min}^{(t)} = \sum_j \alpha_j^{(t)} F_{\min,j}}$. As additional observations arrive, posterior uncertainty over $\theta$ is reduced, producing a tighter estimate of the minimum force needed to lift the object. Thus, $\psi$ refines class-level material priors into instance-level estimates, enabling gentle, material-aware grasping, \textit{in situ}.
\section{Experiments}
\label{sec:results}
\paragraph{Evaluation via Simulations.}
We first compared our approach to open-loop baseline grasp force policies: \textit{Na\"ive}, where the commanded grasp force is always 3~\unit{\newton} (sufficient to lift the largest object in \textit{Squash}) regardless of the object's class, and \textit{Class Prior}, where grasp force is determined using \cref{eq:soft-force-equilibrium} from the class's average material properties in \textit{Squash}. We then compared our approach to adaptive slip-reactive force control methods~\cite{romano2011human} that increase force when incipient slip is detected, and to a direct force predictor that estimates grasp force without estimating material properties. Using the simulator described in \cref{sec:method}, all learned estimators are trained and validated on 200k simulated pick-ups of 85\% of the \textit{Squash} objects, and evaluated on 5k simulations of the remaining 15\%. Both sets use randomly augmented 3D meshes and material properties sampled from the \textit{Squash} dataset, where we procedurally augment meshes to create realistic variations in shape. The ground truth ideal grasp force is determined via a ``drop test'' where applied force (at the evaluated grasp pose) is decreased until gross slip is observed. The results are summarized in \cref{tab:errors}. Our material-adaptive approach outperforms baselines through accurate estimation of the object's material properties (Pearson correlations: $r_E{=}0.65$, $r_m{=}0.88$, $r_\mu{=}0.28$, $r_\tau{=}0.57$), demonstrating its ability to adapt to the specific material properties of each object and maintain a high success rate. Notably, our approach produces an approximately unbiased force estimate, closest to the ideal value of any method, demonstrating that material estimation is beneficial for gentle grasping. We additionally ablate the class prior and the $\tau A$ term of the contact model (\cref{eq:force-hypothesis}) to isolate their contributions. We observe small performance degradations in each case, indicating that material priors are beneficial but not strictly necessary, and that the $\tau A$ term improves force accuracy to the extent outlined in \cref{sec:physics}. The limited degradation from prior ablation also suggests our method may generalize well outside the domain of fruits and vegetables. Additional metrics are in the supplement.

\begin{table}[t]
    \centering
    \scriptsize
    \begin{threeparttable}
        \caption{ 
            \textbf{Efficacy of Material-Aware Robot Grasping.} We compare our material-aware approach for force control with open-loop and closed-loop baselines using 5k simulated objects with high material variability. ``Na\"ive'' uses a force of 3~\unit{\newton} in each gripper jaw, sufficient to lift the largest object in \textit{Squash}. ``Class Prior'' uses the class's average material properties to determine the grasp force using \cref{eq:soft-force-equilibrium}. Direct force prediction estimates the grasp force without intermediate material estimation. Slip-reactive policies modulate grasp force to hover around the boundary where incipient slip is observed. In contrast, our approach refines instance-level material properties for fine-grained force control. The best results are in bold, and second best are underlined.
            \vspace{0.05in}
        }
        \label{tab:errors}
        \setlength{\tabcolsep}{3pt}
        \begin{tabularx}{\linewidth}{@{}Xcccccc@{}}
            \toprule
                                                                             & \multicolumn{3}{c}{Force error (per jaw)} & \multicolumn{3}{c}{Control outcome}                                                                                                                                                                        \\
            \cmidrule(l{\tabcolsep}r{\tabcolsep}){2-4} \cmidrule(l{\tabcolsep}r{\tabcolsep}){5-7}
            Grasp Force Controller                                           & Bias [N]                                  & MAE [N] $\downarrow$                   & WAPE $\downarrow$                      & Excess [N] $\downarrow$                & Deficit [N] $\downarrow$               & Success $\uparrow$                     \\
            \midrule
            \multicolumn{7}{@{}l}{\textit{Open-Loop Baseline Grasping Approaches (Non-Adaptive)}}                                                                                                                                                                                                                                     \\
            \, Na\"ive constant force ($3$~N)                                & 1.97 {\tiny [1.67,\,2.22]}                & 2.18 {\tiny [2.02,\,2.33]}             & 2.11 {\tiny [1.56,\,2.95]}             & 2.07 {\tiny [1.87,\,2.26]}             & \textbf{0.10} {\tiny [0.01,\,0.23]}    & \textbf{1.00} {\tiny [0.99,\,1.00]}    \\
            \, Class prior                                                   & -0.26 {\tiny [-0.57,\,0.10]}              & 0.70 {\tiny [0.50,\,0.95]}             & 0.68 {\tiny [0.59,\,0.77]}             & 0.22 {\tiny [0.14,\,0.31]}             & 0.48 {\tiny [0.27,\,0.74]}             & 0.89 {\tiny [0.77,\,1.00]}             \\
            \multicolumn{7}{@{}l}{\textit{Closed-Loop Baseline Grasping Approaches (Adaptive)\smash{\tnote{a}}}}                                                                                                                                                                                                                      \\
            \, Direct force predictor (learned)                              & 0.29 {\tiny [-0.05,\,0.60]}               & 0.96 {\tiny [0.75,\,1.19]}             & 0.93 {\tiny [0.74,\,1.21]}             & 0.62 {\tiny [0.46,\,0.80]}             & 0.33 {\tiny [0.14,\,0.59]}             & 0.96 {\tiny [0.90,\,1.00]}             \\
            \, Slip-reactive (heuristic, shear)                              & 0.77 {\tiny [0.52,\,0.98]}                & 1.12 {\tiny [0.98,\,1.29]}             & 1.09 {\tiny [0.86,\,1.43]}             & 0.95 {\tiny [0.82,\,1.08]}             & \underline{0.18} {\tiny [0.05,\,0.34]} & 0.96 {\tiny [0.91,\,1.00]}             \\
            \, Slip-reactive (heuristic, contact area)                       & -0.52 {\tiny [-0.82,\,-0.26]}             & 0.71 {\tiny [0.48,\,0.99]}             & 0.69 {\tiny [0.60,\,0.75]}             & \textbf{0.10} {\tiny [0.06,\,0.17]}    & 0.61 {\tiny [0.37,\,0.90]}             & 0.93 {\tiny [0.87,\,1.00]}             \\
            \, Slip-reactive (learned, GBM~\cite{veiga2015stabilizing})      & -0.51 {\tiny [-0.79,\,-0.24]}             & 0.72 {\tiny [0.49,\,1.00]}             & 0.70 {\tiny [0.62,\,0.76]}             & \textbf{0.10} {\tiny [0.07,\,0.13]}    & 0.62 {\tiny [0.38,\,0.91]}             & 0.93 {\tiny [0.86,\,0.99]}             \\
            \, Slip-reactive (learned, images)                               & 0.16 {\tiny [-0.29,\,0.51]}               & 0.88 {\tiny [0.69,\,1.11]}             & 0.85 {\tiny [0.73,\,1.01]}             & 0.47 {\tiny [0.37,\,0.56]}             & 0.41 {\tiny [0.19,\,0.68]}             & 0.97 {\tiny [0.93,\,1.00]}             \\
            \midrule
            \multicolumn{7}{@{}l}{\textbf{\textit{Material-Aware Closed-Loop Grasping Approaches (Adaptive), with Ablations\smash{\tnote{b}}}}}                                                                                                                                                                                       \\
            \, Ours                                                          & \textbf{-0.05} {\tiny [-0.44,\,0.24]}     & \textbf{0.51} {\tiny [0.19,\,0.84]}    & \textbf{0.52} {\tiny [0.43,\,0.60]}    & \underline{0.21} {\tiny [0.16,\,0.26]} & 0.33 {\tiny [0.09,\,0.68]}             & 0.95 {\tiny [0.90,\,0.99]}             \\
            \,\, \rotatebox[origin=c]{180}{$\Lsh$} no prior                  & \underline{-0.14} {\tiny [-0.42,\,0.07]}  & \underline{0.61} {\tiny [0.43,\,0.83]} & \underline{0.59} {\tiny [0.49,\,0.68]} & 0.23 {\tiny [0.17,\,0.29]}             & 0.38 {\tiny [0.19,\,0.62]}             & 0.94 {\tiny [0.84,\,1.00]}             \\
            \,\, \rotatebox[origin=c]{180}{$\Lsh$} wrong (adversarial) prior & -0.20 {\tiny [-0.46,\,0.03]}              & 0.62 {\tiny [0.47,\,0.86]}             & \underline{0.59} {\tiny [0.50,\,0.67]} & \underline{0.21} {\tiny [0.14,\,0.30]} & 0.40 {\tiny [0.21,\,0.64]}             & 0.93 {\tiny [0.82,\,0.97]}             \\
            \,\, \rotatebox[origin=c]{180}{$\Lsh$} Coulomb-only friction     & -0.16 {\tiny [-0.41,\,0.08]}              & 0.62 {\tiny [0.45,\,0.84]}             & 0.61 {\tiny [0.52,\,0.69]}             & 0.24 {\tiny [0.19,\,0.30]}             & 0.38 {\tiny [0.21,\,0.68]}             & 0.94 {\tiny [0.88,\,0.98]}             \\
            \, Ours + safety margin\smash{\tnote{c}}                         & 0.22 {\tiny [-0.14,\,0.59]}               & 1.00 {\tiny [0.77,\,1.26]}             & 0.97 {\tiny [0.78,\,1.24]}             & 0.61 {\tiny [0.40,\,0.86]}             & 0.39 {\tiny [0.20,\,0.61]}             & \underline{0.99} {\tiny [0.96,\,1.00]} \\
            \bottomrule
        \end{tabularx}
        \begin{tablenotes}[flushleft]
            \scriptsize
            \item[a] PR-AUC of slip detection methods: shear (0.05), contact area (0.16), GBM (0.23), images (0.80). Higher is better, random is 0.04.
            \item[b] Ablations: ``no prior'' removes prior conditioning by using a global material distribution; ``wrong prior'' uses a mismatched prior at test time; ``Coulomb-only'' removes the $\tau A$ term from the contact model used to determine required force from estimated material properties.
            \item[c] Commands $\gamma \cdot Q_p(F_\text{min})$ over the posterior; $(\gamma=1.25, p=0.5)$ selected on validation objects.
        \end{tablenotes}
    \end{threeparttable}
    \vspace{-0.3in}
\end{table}

\begin{figure}[p]
    \centering
    \setlength{\tabcolsep}{0.0in}
    \begin{tabular}{@{}cccccccccc}
        \multicolumn{4}{c}{
            \begin{subfigure}{0.4\linewidth}
                \centering
                \includegraphics{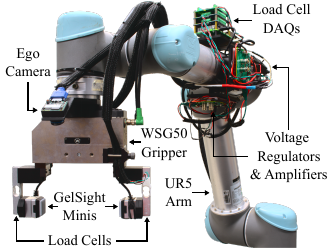}
                \subcaption{Robot Arm and Gripper Hardware}\label{fig:experiments:hardware}
            \end{subfigure}
        }
         &
        \multicolumn{6}{r}{
            \begin{subfigure}{2.94in}
                \centering
                \includegraphics{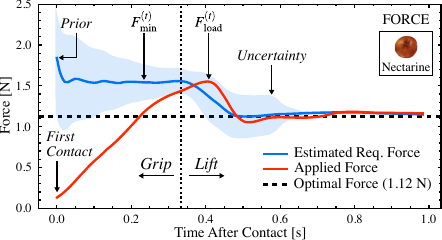}
                \subcaption{Force Evolution Profile During a Grasp}\label{fig:experiments:force-evolution}
            \end{subfigure}
        }
        \\[0.05in]
        \multicolumn{10}{c}{
            \begin{subfigure}{\linewidth}
                \centering
                \includegraphics{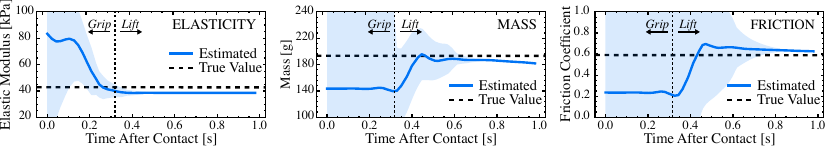}
                \subcaption{Material Property Estimation During a Grasp}\label{fig:experiments:materials-evolution}
            \end{subfigure}
        }
        \\[0.05in]
        \multicolumn{10}{c}{
            \begin{subfigure}{\linewidth}
                \centering
                \includegraphics{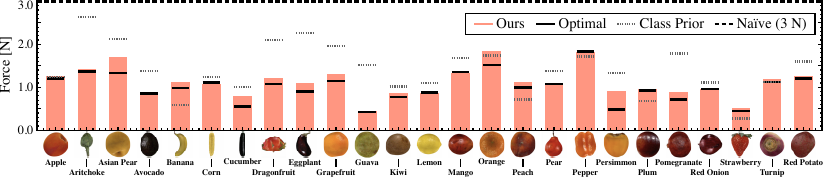}
                \subcaption{Individual Predicted Grasp Forces}\label{fig:experiments:barplot}
            \end{subfigure}
        }
        \\[0.05in]
        \multicolumn{5}{c}{
            \begin{minipage}[b]{0.5\linewidth}
                \centering
                {
                    \scriptsize
                    \setlength{\tabcolsep}{7pt}
                    \def\svgwidth{\linewidth}
                    \begin{tabular}{@{}r}
                        \includegraphics{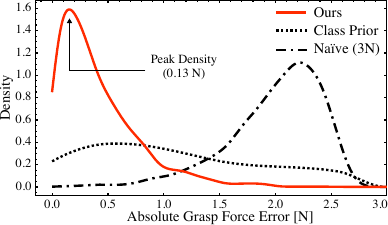}
                        \\ [0.04in]
                        \begin{tabular}{lccc}
                            \toprule
                            Approach                  & MAPE $\downarrow$                          & MAE~[\unit{\newton}] $\downarrow$          & Success $\uparrow$ \\
                            \midrule
                            Na\"ive (3\unit{\newton}) & $2.44 {\scriptstyle \pm 1.49}$             & $1.99 {\scriptstyle \pm 0.33}$             & $\mathbf{1.00}$    \\
                            Class Prior               & \underline{$0.57 {\scriptstyle \pm 0.61}$} & \underline{$0.46 {\scriptstyle \pm 0.40}$} & $0.72$             \\
                            Ours                      & $\mathbf{0.23 {\scriptstyle \pm 0.18}}$    & $\mathbf{0.21 {\scriptstyle \pm 0.14}}$    & \underline{$0.97$} \\
                            \bottomrule
                        \end{tabular}
                    \end{tabular}
                }
                \subcaption{Average Per-Jaw Force Estimation Error}
                \label{fig:experiments:errors}
            \end{minipage}
        }
         &
        \multicolumn{5}{c}{
            \begin{minipage}[b]{0.5\linewidth}
                \centering
                \begin{subfigure}{\linewidth}
                    \centering
                    \includegraphics{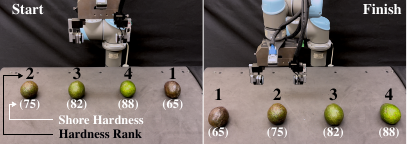}
                    \subcaption{Sorting \textit{Avocados} Based on Ripeness}
                    \label{fig:experiments:sorting-avocado}
                \end{subfigure}

                \vspace{0.075in}

                \begin{subfigure}{\linewidth}
                    \centering
                    \includegraphics{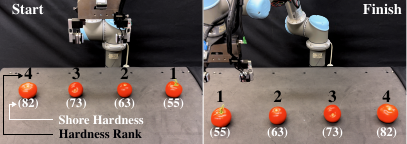}
                    \subcaption{Sorting \textit{Tomatoes} Based on Ripeness}
                    \label{fig:experiments:sorting-tomato}
                \end{subfigure}
            \end{minipage}
        }
    \end{tabular}
    \vspace{-0.05in}
    \caption{
        \textbf{Experimental Results.} (a)~Robot hardware setup for real-world grasping experiments. Our approach is powered by GelSight Mini tactile sensors paired with bending beam load cells, but is not limited to this configuration. (b-c)~Evolution of the applied per-jaw grasp force and estimated object material properties during a trial. The values converge quickly, and, as a result, we gently grasp the object in real time with a force close to the optimal force. (d)~The applied per-jaw grasp force by our approach for various real fruits and vegetables. We consistently apply at or slightly above the optimal force. (e)~Performance of our approach over all 29 tested real fruits and vegetables, where we show accurate force control while maintaining a high grasp success rate. In the kernel density plot, our approach shows a tighter concentration around zero, indicating consistent performance that adapts to individual object properties. (f-g)~Sorting fruits in real time based on their estimated elasticity, which is correlated with ripeness. This is one potential application of material-aware grasping beyond force control.
    }
    \vspace{-0.1in}
    \label{fig:experiments}
\end{figure}

\begin{figure}[t]
    \centering
    \begin{subfigure}{0.315\linewidth}
        \includegraphics[width=\linewidth]{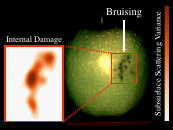}
        \subcaption{Na\"ive (3 N)}
    \end{subfigure}
    \hfill
    \begin{subfigure}{0.315\linewidth}
        \includegraphics[width=\linewidth]{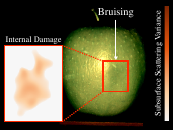}
        \subcaption{Class Prior}
    \end{subfigure}
    \hfill
    \begin{subfigure}{0.315\linewidth}
        \includegraphics[width=\linewidth]{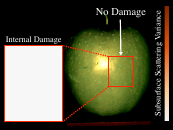}
        \subcaption{Ours}
    \end{subfigure}
    \vspace{-0.05in}
    \caption{
        \textbf{Internal Apple Damage by Grasping Approach.} All images captured using fast separation~\cite{nayar2006fast} immediately after force application, revealing subsurface anomalies invisible to the human eye. High variation in subsurface scattering indicates irregular water content under the apple's skin, which is correlated with bruising. The (a) na\"ive and (b) class prior approaches both leave signs of damage, while (c) our approach does not.
    }\label{fig:fast-separation}
    \vspace{-0.25in}
\end{figure}

\paragraph{Grasping Real Fruits and Vegetables.}
We next assembled a test set of 29 real fruits and vegetables. Their material properties were measured (using the methods in \cref{sec:dataset}) as a ground truth reference. We then performed 3 grasping trials per object with our material-aware approach, each utilizing a different grasp pose, using the robot hardware shown in \cref{fig:experiments:hardware}. After completing each grasping trial, at the executed grasp pose, the applied force is reduced until gross slip is observed to empirically determine the minimum required grasp force for the object. In \cref{fig:experiments:force-evolution,fig:experiments:materials-evolution}, we show the evolution of the applied grasp force and estimated material properties over the duration of grasping a nectarine. As the robot progresses through the \textit{grip} phase, the estimate of elasticity is refined. During the \textit{lift} phase, mass and friction parameters are updated. The force and material estimates converged to stable values quickly over the course of the grasp, enabling the robot to apply a force that never significantly exceeds the optimal force for the nectarine. \cref{fig:experiments:barplot} compares the grasp forces of our method with those of the \textit{Na\"ive} and \textit{Class Prior} baselines, for a wide variety of fruits and vegetables.
\begin{wrapfigure}{r}{1.5in}
    \vspace{-0.05in}
    \centering
    \def\svgwidth{\linewidth}
    {
        \scriptsize
\begingroup%
  \makeatletter%
  \providecommand\color[2][]{%
    \errmessage{(Inkscape) Color is used for the text in Inkscape, but the package 'color.sty' is not loaded}%
    \renewcommand\color[2][]{}%
  }%
  \providecommand\transparent[1]{%
    \errmessage{(Inkscape) Transparency is used (non-zero) for the text in Inkscape, but the package 'transparent.sty' is not loaded}%
    \renewcommand\transparent[1]{}%
  }%
  \providecommand\rotatebox[2]{#2}%
  \newcommand*\fsize{\dimexpr\f@size pt\relax}%
  \newcommand*\lineheight[1]{\fontsize{\fsize}{#1\fsize}\selectfont}%
  \ifx\svgwidth\undefined%
    \setlength{\unitlength}{109.83641624bp}%
    \ifx\svgscale\undefined%
      \relax%
    \else%
      \setlength{\unitlength}{\unitlength * \real{\svgscale}}%
    \fi%
  \else%
    \setlength{\unitlength}{\svgwidth}%
  \fi%
  \global\let\svgwidth\undefined%
  \global\let\svgscale\undefined%
  \makeatother%
  \begin{picture}(1,1.1471602)%
    \lineheight{1}%
    \setlength\tabcolsep{0pt}%
    \put(0,0){\includegraphics[width=\unitlength,page=1]{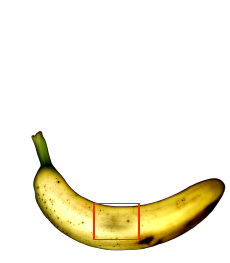}}%
    \put(0.50835903,0.29626519){\color[rgb]{1,0.16862745,0}\makebox(0,0)[t]{\lineheight{1.03748071}\smash{\begin{tabular}[t]{c}Bruising\end{tabular}}}}%
    \put(0.04403217,0.26220812){\color[rgb]{0,0,0}\rotatebox{90}{\makebox(0,0)[t]{\lineheight{1.03748071}\smash{\begin{tabular}[t]{c}\textbf{Class Prior}\end{tabular}}}}}%
    \put(0.04403217,0.75384834){\color[rgb]{0,0,0}\rotatebox{90}{\makebox(0,0)[t]{\lineheight{1.03748071}\smash{\begin{tabular}[t]{c}\textbf{Ours}\end{tabular}}}}}%
    \put(0,0){\includegraphics[width=\unitlength,page=2]{banana-damage.pdf}}%
  \end{picture}%
\endgroup%

    }
    \vspace{-0.175in}
    \caption{
        \textbf{Surface Damage After 1 Day.}
        Contrast-enhanced images 1 day after grasp force application. The force predicted by our approach leaves no visible damage. The class prior approach overexerts, leaving a bruise.
    }\label{fig:banana-damage}
    \vspace{-0.4in}
\end{wrapfigure}
As seen in \cref{fig:experiments:errors}, the distribution of excess forces for our approach is more tightly concentrated around zero, compared to the baselines. Finally, in \cref{fig:experiments:sorting-avocado,fig:experiments:sorting-tomato}, we show how online material estimation can enable higher-order planning. As ripeness is correlated with hardness in climacteric fruits, by estimating the elasticity of each fruit during its grasp, the robot can sort them based on their ripeness in real time.

\paragraph{Qualitative Damage Evaluation.}
Since applying excess force increases the risk of damaging the object being grasped, we have thus far used excess force as a quantitative proxy for damage. However, using fast separation~\cite{nayar2006fast}, we can image internal bruising from each grasping method. In \cref{fig:fast-separation}, we project high-frequency illumination on an apple immediately after completing a grasp. From local variations in subsurface scattering, we can then reveal any hidden damage. Further, in \cref{fig:banana-damage}, we can see surface damage to the banana 1 day after being grasped. In both cases, only our approach shows no visible damage.

\section{Limitations and Directions for Future Work}
\label{sec:limitations}
We have taken a first step towards robot grasping that uses \textit{just enough} force to handle an object. In our work, we focused on the task of minimizing the grasp force given a suitable grasp pose for a parallel-jaw gripper. To minimize the potential for damage, we would also need to address the choice of grasp poses that do not require large forces (\eg, a banana picked up by one end). While we have used a parallel-jaw gripper, further analysis of the physics of grasping would be needed to extend our work to more complex end-effectors, such as dextrous hands that can cradle objects. Additionally, the performance of our method is tied to that of the tactile sensor we use. Current sensors trade off spatial and temporal resolution. While we addressed this problem by using interpolation, to improve performance we will need sensors with both high update rate and spatial resolution. Furthermore, we have focused on the domain of fruits and vegetables, for which we could measure class priors. For objects without such priors, our method may still be effective, but this requires additional validation. Finally, we have assumed that the object's material properties are constant over its surface. We plan to extend our work to the harder problem of grasping objects with spatially varying properties.
\section{Conclusion}
\label{sec:conclusion}
We have introduced material-aware robot grasping: a visuotactile approach that estimates instance-level material properties, online, during contact with an object. This allows us to drive closed-loop force control towards the theoretical minimum grasp force. We created \textit{Squash}, a diverse dataset of 3D object models with measured material properties. Our visuotactile model, trained using \textit{Squash}, recovers elasticity, mass, and friction parameters from tactile feedback. In simulation and on real hardware, our method consistently achieved gentle, non-damaging grasps. In addition, the \textit{in situ} material estimation of our method opens avenues for object sorting, grasp-pose optimization, dexterous manipulation, or any task where knowing \emph{how} an object deforms is of value.

\clearpage
\acknowledgments{
    This work was supported in part by the Office of Naval Research (ONR) award N00014-24-1-2155 and in part by the National Science Foundation (NSF) and Center for Smart Streetscapes (CS3) under NSF Cooperative Agreement No.\ EEC-2133516. The authors are grateful to Behzad Kamgar-Parsi at ONR for his support. The authors thank Dhruv Yalamanchi and Isabel Tu for help with collecting the \textit{Squash} dataset, Joaquin Palacios for discussions on robot control, and Silvia Sell\'an for visualization support.
}

\bibliography{main}  %

\clearpage

\renewcommand{\thesection}{S\arabic{section}}
\renewcommand{\thefigure}{S\arabic{figure}}
\renewcommand{\thetable}{S\arabic{table}}
\renewcommand{\theequation}{S\arabic{equation}}
\setcounter{section}{0}
\setcounter{figure}{0}
\setcounter{table}{0}
\setcounter{equation}{0}
\setcounter{footnote}{0}

\maketitlesupplementary

\section{Methods Used to Collect the \textit{Squash} Dataset}

To capture the geometry and texture of each object, we used a Shining 3D Einscan SP V2 structured light scanner which has a single shot accuracy of ${\leq}0.05\,\text{mm}$. The objects are first placed in an initial orientation on the turntable, and rotated to capture the full surface using between 8--36 views, depending on the texture of the object. The object is then reoriented at least one additional time, and the process is repeated to ensure comprehensive coverage. The resulting 3D meshes are postprocessed to be watertight, and have high resolution to capture the fine details in the object's shape and texture. The 3D scan additionally provides a measurement of the object's volume, and by weighing the object with a kitchen scale to get its mass, we can calculate its density.

\begin{figure}[t]
    \centering
    \begin{subfigure}[b]{\linewidth}
        \centering
        {
            \scriptsize
            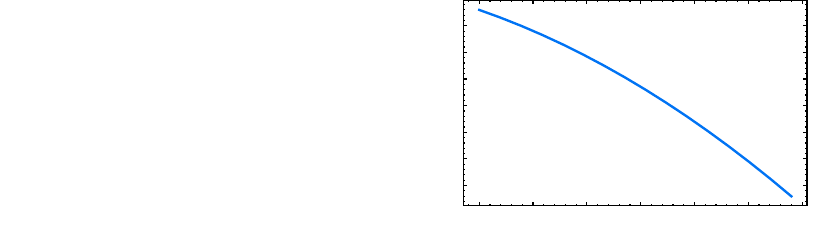
        }
        \subcaption{Compression Test}\label{supp:fig:measurement:elasticity}
        \vspace{0.05in}
    \end{subfigure}
    \begin{subfigure}[b]{\linewidth}
        \centering
        {
            \scriptsize
            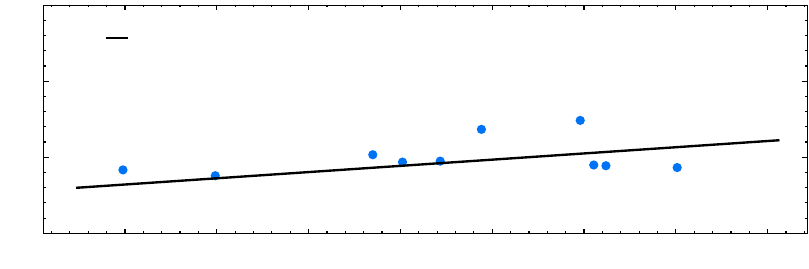
        }
        \vspace{0.05in}
        \subcaption{Tribological Test}\label{supp:fig:measurement:friction-shear}
    \end{subfigure}
    \caption{
        \textbf{Extracting Material Properties from Probing Tests.} The above plots show the probing results for a single instance of a peach. (a)~A compression test is performed to measure the elastic modulus of the object, which is reported as a function of the Poisson ratio. (b)~A tribological test is performed to measure the friction coefficient and shear strength of the object, where each point corresponds to a different trial at a different location on the object's surface. The fitted line is used to decouple the two properties during estimation.
    }\label{supp:fig:measurement}
    \vspace{-0.2in}
\end{figure}

To measure the object's elastic modulus, we perform a uniaxial compression test with a parallel-jaw gripper. As force is applied to the object, the resulting deformation is observed by tactile sensors at the jaw fingertips. The relationship between the applied force and the resulting deformation is used to calculate the effective elastic modulus of both the tactile sensor (GelSight gel pad) and object. As shown in \cref{supp:fig:measurement:elasticity}, we have several measurements of the effective modulus during closure, which we average for a final estimate. The object's elastic modulus is then recovered through calibration of the tactile fingertip's own elastic modulus, \ie, by using a rigid standard object. The object's elastic modulus is reported as a function of an assumed Poisson ratio (\cref{supp:fig:measurement:elasticity}), since we can not directly measure it using our robotic setup.

To measure the object's (static) surface friction $\mu$ and interfacial shear strength $\tau$, we perform a tribological ``drop test''. The object starts lifted and the applied force is reduced until the object slips, which is detected through optical flow in the tactile sensor. The applied force and contact area at the moment of slip are then used to calculate $\mu$ and $\tau$ through the soft friction model in \cref{eq:soft-force-equilibrium}. Several trials are performed per object, each at a different grasp location. As seen in \cref{supp:fig:measurement:friction-shear}, the multiple samples allow us to fit a linear model to recover $\mu$ and $\tau$ during estimation.

\section{Deformable Grasp Simulation Details}
\label{supp:sec:simulation}

Below we provide additional details on our deformable grasp simulation, including the physics engine, scene configuration, control strategy for data generation, and the data types that are captured during the course of a simulated grasp. We use this simulation to generate 200k training grasp sequences for our visuotactile model $\psi$, which estimates the material properties of the grasped object.

\subsection{Physics Engine and Constitutive Model}
We simulate robotic grasps using UIPC~\cite{huangGIPCFastStable2024,huangStiffGIPCAdvancingGPU2025}, a GPU-accelerated Incremental Potential Contact (IPC) solver~\cite{liIncrementalPotentialContact2020}, which has been modified to additionally model the effects of adhesion. In \textit{Squash}, we have triangular surface meshes captured through structured light scanning. We first ensure each surface mesh is manifold and non-self-intersecting, then tetrahedralise the surface meshes prior to simulation. Each simulated grasped object is treated as a deformable body, modeled with a Stable Neo-Hookean constitutive law, and discretized using first-order tetrahedral finite elements (FEM). In our case, since our target real-world robot platform uses GelSight tactile sensors as the gripper fingertips, we also model the gripper fingertips (gel pads) as deformable bodies in the same manner. Contact is resolved with IPC's barrier-based formulation, which guarantees intersection-free trajectories throughout the simulation. The contact distance threshold is set to $\hat{d}{=}1\,\text{mm}$, and a simulation time step of $\Delta t{=}4\,\text{\unit{\milli\second}}$ is used throughout.

\subsection{Scene Configuration}
Each simulation scene contains three bodies: a left gripper finger, a right gripper finger, and a grasped object. The gripper fingers are high-resolution meshes of the GelSight Mini gel pad. They are assigned an elastic modulus of $E_\text{grip}{=}245\,\text{\unit{\kilo\pascal}}$, Poisson ratio $\nu_\text{grip}{=}0.48$, mass $m_\text{grip}{=}3\,\text{\unit{\gram}}$, friction coefficient $\mu_\text{grip}{=}0.9$, and interfacial shear strength $\tau_\text{grip}{=}100\,\text{\unit{\kilo\pascal}}$, chosen to match the silicone elastomer of the physical sensor.

The grasped object is assigned material properties $(E, m, \mu, \tau)$ that constitute the estimation target of the learned model. These properties are drawn from per-class Gaussian distributions fitted to the empirical property measurements in \textit{Squash}. The Gaussian distributions are truncated to physically valid values. Poisson ratio $\nu$ is not directly measured in \textit{Squash}; instead we fit a curve which models the relationship between elastic modulus $E$ and an assumed value for $\nu$. We thus assume all objects have $\nu{=}0.48$, and adjust our value for $E$ accordingly. Inter-object friction and contact compliance are computed with the harmonic mean and the Hertz effective modulus, respectively, ensuring physical consistency at each contact pair.

To maximize coverage of contact geometry, each object is randomly oriented before placement. Each Euler angle is sampled independently from $[0,\,2\pi)$, and the object is subsequently aligned so that a randomly chosen in-plane direction (lying in the $yz$-plane) coincides with the object's principal geometric axis. In addition, the object shape is augmented to vary its dimensions. The object is then centered at the scene origin and the gel pads are positioned symmetrically about it along the $x$-axis, with an initial gap just beyond the contact distance threshold $\hat{d}$.

\subsection{Simulated Gripper Control}
\begin{figure}
    \centering
    {
        \footnotesize
\begingroup%
  \makeatletter%
  \providecommand\color[2][]{%
    \errmessage{(Inkscape) Color is used for the text in Inkscape, but the package 'color.sty' is not loaded}%
    \renewcommand\color[2][]{}%
  }%
  \providecommand\transparent[1]{%
    \errmessage{(Inkscape) Transparency is used (non-zero) for the text in Inkscape, but the package 'transparent.sty' is not loaded}%
    \renewcommand\transparent[1]{}%
  }%
  \providecommand\rotatebox[2]{#2}%
  \newcommand*\fsize{\dimexpr\f@size pt\relax}%
  \newcommand*\lineheight[1]{\fontsize{\fsize}{#1\fsize}\selectfont}%
  \ifx\svgwidth\undefined%
    \setlength{\unitlength}{396bp}%
    \ifx\svgscale\undefined%
      \relax%
    \else%
      \setlength{\unitlength}{\unitlength * \real{\svgscale}}%
    \fi%
  \else%
    \setlength{\unitlength}{\svgwidth}%
  \fi%
  \global\let\svgwidth\undefined%
  \global\let\svgscale\undefined%
  \makeatother%
  \begin{picture}(1,0.31222775)%
    \lineheight{1}%
    \setlength\tabcolsep{0pt}%
    \put(0,0){\includegraphics[width=\unitlength,page=1]{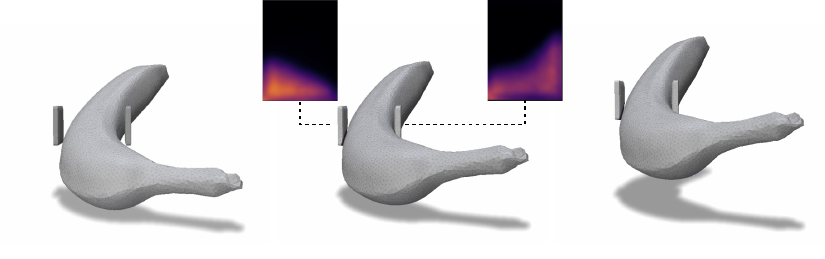}}%
    \put(0.16450935,0.00417256){\color[rgb]{0,0,0}\makebox(0,0)[t]{\lineheight{0.80610013}\smash{\begin{tabular}[t]{c}\textbf{\textit{Approach}}\end{tabular}}}}%
    \put(0.50087297,0.00417256){\color[rgb]{0,0,0}\makebox(0,0)[t]{\lineheight{0.80610013}\smash{\begin{tabular}[t]{c}\textbf{\textit{Grip}}\end{tabular}}}}%
    \put(0.83723659,0.00417256){\color[rgb]{0,0,0}\makebox(0,0)[t]{\lineheight{0.80610013}\smash{\begin{tabular}[t]{c}\textbf{\textit{Lift}}\end{tabular}}}}%
    \put(0.36363636,0.28192471){\color[rgb]{1,1,1}\makebox(0,0)[t]{\lineheight{0.80610013}\smash{\begin{tabular}[t]{c}$P_{\text{left}}^{(t)}$\end{tabular}}}}%
    \put(0.63636364,0.28192471){\color[rgb]{1,1,1}\makebox(0,0)[t]{\lineheight{0.80610013}\smash{\begin{tabular}[t]{c}$P_{\text{right}}^{(t)}$\end{tabular}}}}%
  \end{picture}%
\endgroup%

    }
    \vspace{-0.15in}
    \caption{\textbf{Simulating Deformable Object Grasps.} We use UIPC, a GPU-accelerated incremental potential contact solver, to simulate the dynamics of a parallel-jaw gripper with deformable elastomeric gel pads grasping a deformable object. Deformable objects are taken from the \textit{Squash} dataset, and their material properties and dimensions are augmented to vary the contact geometry. The simulator extracts rich visuotactile observations at each time step, such as the pressure map $\smash{P^{(t)}}$ observed in the local frame of the jaw pads, which are used to train our learned model for material estimation and force control.}
    \label{supp:fig:simulated-grasp}
    \vspace{-0.2in}
\end{figure}

We employ an \emph{adaptive force-controlled} gripper. During the \textit{approach} phase, both jaws close at a constant speed of $v_\text{close}{=}4\,\text{\unit{\milli\meter\per\second}}$ until the contact force on either jaw exceeds a contact detection threshold of $0.25\,\text{\unit{\newton}}$, or until the jaw displacement reaches 30\% of the object's half-width plus $\hat{d}$ (whichever comes first). Once contact is detected (\ie, the \textit{grip} phase has begun), the jaws enter a proportional (P) control loop that adjusts their displacement to drive the measured contact force $F_\text{load}$ toward a target grip force $F^*$. The control law is given by:
\begin{align}
    x \leftarrow x
    + K_p\,(F^* - F_\text{load})\,\text{sgn}(v_\text{close}),
    \quad K_p = 10^{-4}\,\text{m\,N}^{-1}.
\end{align}
Transition to the \textit{lift} phase is triggered once both jaws have been simultaneously in servo for at least five time-steps \emph{and} the force error is within 20\% of $F^*$, or after a maximum of 100 servo steps. After a brief $100\,\text{\si{ms}}$ dwell, the gripper lifts at
$v_\text{lift}{=}10\,\text{mm\,s}^{-1}$ over a distance of $2\,\text{mm}$. This process is outlined in \cref{supp:fig:simulated-grasp}.

\subsection{Sensor Observation}
At every time step, the simulator extracts the following tactile observations from each gripper jaw:

\begin{itemize}
    \item \textbf{Scalar contact force} $F_\text{load} \in \mathbb{R}_{\geq 0}$ [\si{N}]: the $L^2$ norm of the summed contact gradient vectors over all contact vertices, divided by $\Delta t^2$.
    \item \textbf{Force vector} $\mathbf{F}_\text{load} \in \mathbb{R}^3$ [\si{N}]: the vector sum of per-vertex contact forces in the gripper jaw's local coordinate frame.
    \item \textbf{Total contact area} $a$ [\si{m^2}]: the total area of surface triangles on the gripper jaw pad that contain at least one vertex in contact with the object.
    \item \textbf{Contact centroid} $\mathbf{c} \in \mathbb{R}^3$: the mean position of contact vertices, normalized by the bounding box of the gripper jaw pad and expressed relative to the pad center.
    \item \textbf{Gripper displacement} $\mathbf{d} \in \mathbb{R}^3$ [\si{m}]: the rigid-body displacement of the pad from its initial rest position at the beginning of the simulated grasp.
    \item \textbf{Force map (tactile image)} $I \in \mathbb{R}^{H \times W}$ [\si{N}]: an image in which each pixel accumulates the contact force contribution of overlapping surface triangles, rasterized in the gripper jaw pad's $yz$-plane.
    \item \textbf{Pressure map} $P \in \mathbb{R}^{H \times W}$ [\si{Pa}]: per-vertex contact pressure (area-weighted average of incident triangle pressures), rendered with barycentric interpolation into the same image space as the force map. An example of the pressure map is shown in \cref{supp:fig:simulated-grasp}.
    \item \textbf{Indentation map} $L \in \mathbb{R}^{H \times W}$ [\si{m}]: per-vertex surface indentation depth, computed as the positive part of the difference between rest and deformed $z$-coordinates in the pad's body frame, and rendered with barycentric interpolation.
    \item \textbf{Contact area map} $A \in \{0,1\}^{H \times W}$: binary mask indicating pixels with non-zero force.
    \item \textbf{Shear map} $V \in \mathbb{R}^{H \times W}$ [\si{m}]: per-vertex shear displacement, computed as the $L^2$ norm of the $yz$-components of the vertex displacement, and rendered with barycentric interpolation.
    \item \textbf{Slip map} $U \in \mathbb{R}^{H \times W}$ [\si{m}]: per-vertex slip displacement, computed as the difference between the current and previous time step's contact centroids, and rendered with barycentric interpolation.
\end{itemize}

The complete time-series of these observations, together with the object's mechanical properties and grip force used in that trial, is stored as a compressed NumPy archive (\texttt{.npz}).

\subsection{Parallel and Distributed Execution}

To amortize GPU setup overhead, we batch multiple grasps into a single UIPC scene by tiling simulations on a 2D grid in the $xz$-plane. A data manifest specifies all combinations of object mesh, material properties, and grip force to be simulated; this manifest is partitioned into contiguous chunks and each chunk is dispatched to a separate GPU subprocess, enabling large-scale parallel data collection across multi-GPU compute nodes.

\subsection{IPC Simulation Sensitivity}

To evaluate our simulator's fidelity for grasping, we ran a sensitivity study over the IPC solver parameters. While per-object target grasp forces shift modestly ($\sim$$0.1$~N), our method's MAE moves by $\leq$$5\%$. Force estimates are insensitive to loosening friction regularization ($\epsilon_v$$\times$$10$: -$5\%$), halving/doubling the contact barrier (-$4\%$), and halving the time step $dt$ (+$3\%$). Larger $\epsilon_v$ or $dt$ tend to reduce $\mu$ and $\tau$ identifiability, as incipient slip becomes less apparent.

\section{Visuotactile Model Design and Training Details}

We now discuss the architecture of our learned model $\psi$ for estimating material properties from visuotactile sequences, as well as details of its training and deployment. The model consists of a sequence encoder that processes sequential tactile observations and control signals, a likelihood head that scores candidate material parameter vectors, and an online importance-sampling estimator that maintains a weighted belief over the material parameters.

\subsection{Model Architecture}

\paragraph{Feature Representation.}
We represent the material properties of each object as a physical parameter vector $\theta = (E, m, \mu, \tau)$, where $E$ is elastic modulus, $m$ is mass, $\mu$ is the Coulomb friction coefficient, and $\tau$ is the interfacial shear strength. For inference, we work in a log-parameter space:
\begin{align}
    \mathbf{z} = \bigl[\ln E,\;\ln m,\;\ln \mu,\;\ln \tau\bigr] \in \mathbb{R}^4,
\end{align}
which maps the physically constrained parameters onto an unconstrained real-valued vector suitable for likelihood evaluation and importance sampling. The inverse mapping $\mathbf{z} \to \theta$ applies elementwise exponentials. We evaluate the likelihood in this log-parameter space, with each $\bm{z}$ corresponding uniquely to the physical parameters $\theta=g^{-1}(\bm{z})$.

At each time step $t$ the robot's tactile sensors (one per jaw) yield a tactile image $\smash{I^{(t)}} \in \mathbb{R}^{H \times W}$. Each jaw's image is encoded by a shared convolutional backbone: three strided Conv2d layers (channels $1{\to}16{\to}32{\to}64$, kernel sizes 5/3/3, stride 2) followed by adaptive average pooling and a linear projection to an embedding of dimension $d_\text{emb}{=}128$. The two jaw embeddings are concatenated and fused by a single linear layer to produce a per-timestep feature vector $\mathbf{x}_\text{feat}^{(t)} \in \mathbb{R}^{d_\text{emb}}$. We then assemble a 28-dimensional control vector comprising raw signals (per-jaw normal force, jaw displacement in 3D, contact area, mean contact pressure, and contact centroid in 3D; 18 dimensions total) and their first-order temporal differences (force, area, and centroid; 10 dimensions). These are projected to $\mathbb{R}^{d_\text{emb}}$ by a two-layer MLP to yield $\mathbf{x}_\text{ctrl}^{(t)}$.

\paragraph{Sequence Encoder.}
The concatenated feature and control embedding $\mathbf{x}^{(t)} = [\mathbf{x}_\text{feat}^{(t)};\, \mathbf{x}_\text{ctrl}^{(t)}] \in \mathbb{R}^{2d_\text{emb}}$ is processed by a Closed-form Continuous-time (CfC) recurrent network~\cite{hasaniClosedformContinuoustimeNeural2022}. Concretely, given the sequence $(\mathbf{x}^{(1)}, \ldots, \mathbf{x}^{(t)})$, the CfC produces hidden states:
\begin{align}
    \bm{h}^{(t)} = \mathrm{CfC} \bigl(\mathbf{x}^{(t)},\, \bm{h}^{(t-1)} \bigr),
    \quad \bm{h}^{(t)} \in \mathbb{R}^{d_h},
\end{align}
with hidden dimension $d_h{=}256$. The network is strictly causal: $\bm{h}^{(t)}$ depends only on past and present observations, enabling online inference.

\paragraph{Likelihood Head.}
The hidden state $\bm{h}^{(t)}$ is used to score candidate parameter vectors via a learned likelihood head. A candidate $\tilde{\mathbf{z}} \in \mathbb{R}^4$ is first projected to $\mathbb{R}^{d_\text{emb}}$ by a two-layer MLP. The projected candidate and the hidden state are concatenated and passed through a two-layer scoring-head MLP. The MLP produces a raw compatibility logit, which is scaled by the learned temperature $\beta$; we denote the resulting unnormalized cumulative log-likelihood score by $S_\psi$:
\begin{align}
    S_\psi (\bm{h}^{(t)}, \tilde{\mathbf{z}}) = \log p_\psi(I^{(0:t)}, F_{\text{load}}^{(0:t)} \mid \tilde{\mathbf{z}}) + C^{(t)}. 
\end{align}
Because \(C^{(t)}\) is independent of the candidate parameters, it cancels when the importance weights are normalized. 

\paragraph{Online Importance Sampling.}
At the start of each grasp, we draw $N{=}512$ material samples $\theta_j \sim p_c(\theta)$ and transform each to its log-parameter space representation $\bm{z}_j = g(\theta_j)$ via the log transform above. These samples are held fixed throughout the grasp. At each timestep, the likelihood head updates only their normalized importance weights as follows:
\begin{enumerate}
    \item \textbf{Score.} For each fixed sample $\bm{z}_j$, the likelihood head evaluates its cumulative log-likelihood using the current hidden state,
    \begin{align}
        S_j^{(t)} = S_\psi(\bm{h}^{(t)}, \bm{z}_j).
    \end{align}
    All $N$ samples are evaluated in a single batched forward pass.

    \item \textbf{Reweight.} Log-sum-exp normalization yields importance weights:
    \begin{align}
        \alpha_j^{(t)} = \frac{\exp (S_j^{(t)})}{\sum_k \exp (S_k^{(t)})}.
    \end{align}
    Since the samples are drawn from the prior $p_c(\theta)$, which also serves as the proposal distribution, the prior/proposal terms cancel in the self-normalized importance weights.
\end{enumerate}

Each sample corresponds to the physical parameters $\theta_j$, which are mapped through the grasp model:
\begin{align}
    f(\theta) = \min\nolimits_{F \ge 0} \left\{ F : \mu F + \tau A(F;E) \ge mg \right\}
\end{align}
to obtain \(F_{\min,j}=f(\theta_j)\). The controller uses the posterior expected minimum force,
\begin{align}
    F_{\min}^{(t)} = \sum_j \alpha_j^{(t)} F_{\min,j},
\end{align}
as the target grasp force. For reporting material estimates, we use the posterior mean:
\begin{align}
    \theta^{(t)} = \sum_j \alpha_j^{(t)} \theta_j.
\end{align}

\subsection{Model Training from Precomputed Simulations}

The network is trained on 200k simulated grasping sequences with ground-truth material parameters, as described in \cref{sec:method} of the main paper and \cref{supp:sec:simulation} above. Each sequence consists of $T$ timesteps of tactile images and applied loads paired with a fixed ground-truth $\mathbf{z}_\text{gt}$.

We use a hybrid contrastive-regression objective. The primary term is an InfoNCE loss: at each time step $t$, the hidden state $\bm{h}^{(t)}$ scores one positive ($\mathbf{z}_\text{gt}$) against $K{=}127$ negatives sampled from the class-conditioned prior $p_c(\theta)$, plus all other in-batch ground-truth vectors as cross-negatives. Prior-sampled negatives encourage the contrastive score to approximate the cumulative log-likelihood up to a candidate-independent term; in-batch cross-negatives provide additional discriminative supervision. Let $S^+$ and $S_k^-$ denote the likelihood-head scores of the positive and negative candidates. The loss for batch element $b$ at timestep $t$ is then:
\begin{align}
    \mathcal{L}^{\text{NCE}}_{b,t} = -\log \left ( \frac{\exp(S^+)}
    {\exp(S^+) + \sum_{k} \exp(S_k^-)} \right ).
\end{align}
The scores $S^+$ and $S_k^-$ include the learned temperature scaling $\beta$ defined above. A secondary MSE regression term directly predicts $\mathbf{z}_\text{gt}$ from $\bm{h}^{(t)}$, weighted by $\lambda{=}0.1$. Both terms are averaged using a binary mask $\mathbf{v}_{b,t} \in \{0,1\}$:
\begin{align}
    \mathcal{L} = \frac{\sum_{b,t} \mathbf{v}_{b,t}\bigl(\mathcal{L}^{\text{NCE}}_{b,t}
        + \lambda\,\mathcal{L}^{\text{reg}}_{b,t}\bigr)}{\sum_{b,t} \mathbf{v}_{b,t}}.
\end{align}

The network is trained with Adam ($\text{lr}{=}10^{-4}$) for 500 epochs with a batch size of 8 sequences using a cosine learning rate schedule. Training is distributed across 8 NVIDIA RTX 6000 Ada Generation GPUs, taking approximately 48 hours to complete.

\section{Additional Experimental Results}

Below, we provide additional evaluations of our approach when grasping simulated and real objects.

\paragraph{Error Metrics.} We measure the performance of our visuotactile model using several metrics. Bias measures, in Newtons, whether the approach tends to undershoot or overshoot. Mean absolute error (MAE) measures the absolute error in Newtons. Weighted absolute percentage error (WAPE) and mean absolute percentage error (MAPE) measure performance on a normalized scale. Overload reports the peak force exerted above the optimal force, and is in essence a measure of overshoot. Excess work and excess impulse convey the risk of damage to the object. Success rate is the percentage of grasps where the object did not fall from the grasp of the gripper after it has been lifted.

\begin{figure}[t]
    \centering
    \begin{subfigure}{\linewidth}
        \def\svgwidth{\linewidth}
        {
            \scriptsize
            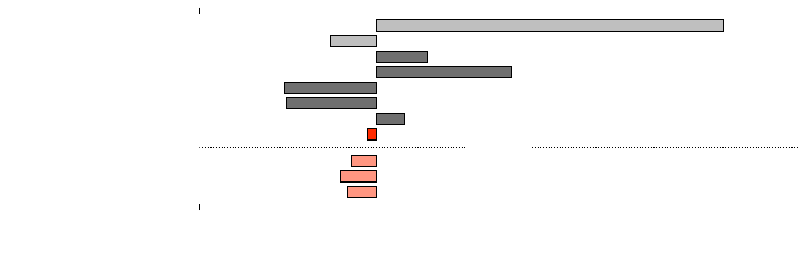
        }
        \vspace{-0.15in}
    \end{subfigure}
    \begin{subfigure}{\linewidth}
        \def\svgwidth{\linewidth}
        {
            \scriptsize
            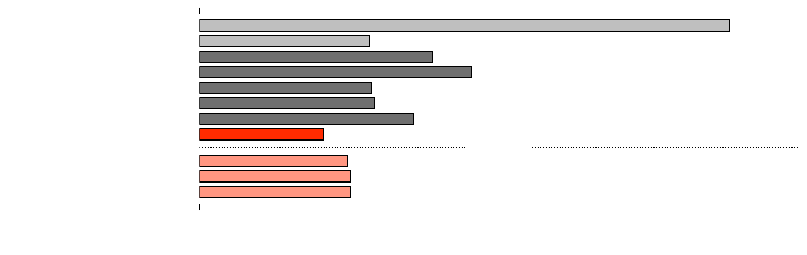
        }
        \vspace{-0.15in}
    \end{subfigure}
    \begin{subfigure}{\linewidth}
        \def\svgwidth{\linewidth}
        {
            \scriptsize
            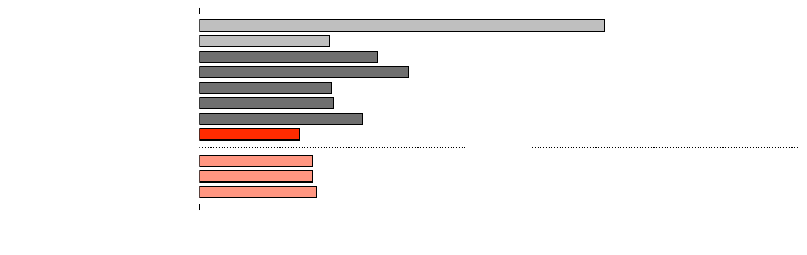
        }
        \vspace{-0.15in}
    \end{subfigure}
    \vspace{-0.25in}
    \caption{\textbf{Force Estimation Error on Simulated Object Grasps.} The bar plots show the (top) signed error/bias, (middle) mean absolute error, and (bottom) weighted absolute percentage error of the estimated minimum required grip force $F_{\min}$ compared to the ground truth, across all simulated grasping sequences. Error bars show the confidence interval. The error is computed once the object has been lifted by 2~\unit{\milli\meter}, which is 0.2~\unit{\second} after lifting begins. The best approach is indicated by a $\star$. The results demonstrate that our visuotactile model is able to accurately estimate the required grip force for a wide range of deformable objects with varying material properties.}\label{supp:fig:simulated-force-error-bars}
    \vspace{-0.01in}
\end{figure}

\paragraph{Results and Ablations in Simulation.} In our simulation experiments, our visuotactile model achieved a high success rate in predicting optimal grasp forces for a wide range of deformable objects. This is seen in \cref{supp:fig:simulated-force-error-bars}, where, on average, our approach exerts a much lower force than baseline methods to successfully complete a grasp. The visuotactile model was able to adapt its predictions based on the material properties of the objects, such as their elasticity, mass, and friction coefficients, as shown in \cref{supp:tab:simulated-errors}. The results showed that our model outperformed baseline methods that did not utilize tactile feedback, demonstrating the importance of incorporating tactile information for gentle grasping of deformable objects. In 83.0\% of cases, our approach yielded the most accurate grasp force estimate, compared to 15.4\% for the \textit{Class Prior} approach and 1.6\% for the \textit{Na\"ive} approach. In addition to our stated method of refining class-level material priors to instance-level posteriors, we perform two ablation tests. In the first, we removed the class-level material prior entirely. In the second, we set the class prior to an incorrect distribution for the object being grasped (\eg, using the prior for apples when grasping an apricot).  We observed that our approach maintains the ability to reasonably estimate material properties from tactile feedback, and thus achieve gentle grasps in both ablation tests.

\begin{table}[t]
    \centering
    \scriptsize
    \begin{threeparttable}
        \caption{ \textbf{Efficacy of Material-Aware Robot Grasping in Simulation (Continuation of \cref{tab:errors}).} We compare our material-aware approach for gripper force control with two standard approaches using 5k simulated produce items with high variability. ``Na\"ive'' uses a force of 3~\unit{\newton} in each gripper jaw, which just exceeds the threshold to successfully lift the largest object in \textit{Squash}. ``Class Prior'' uses the class's average material properties to determine the grasp force using \cref{eq:soft-force-equilibrium}. In contrast, our approach refines instance-level material properties to yield improved fine-grained force control. We additionally report the performance of our approach when removing the class prior (from \textit{observation}), and when the class prior is incorrectly identified. We see that even without an accurate estimate of the object's class-level material prior, our approach achieves gentle grasps relative to the non-adaptive baselines. The best results are in bold, and second best are underlined.
            \vspace{0.05in}
        }
        \label{supp:tab:simulated-errors}
        \setlength{\tabcolsep}{3pt}
        \begin{tabularx}{\linewidth}{@{}Xcccccccc@{}}
            \toprule
                                                                             & \multicolumn{4}{c}{Material estimation MAPE $\downarrow$} & \multicolumn{3}{c}{Force error $\downarrow$}                                                                                                                                                                                          \\
            \cmidrule(l{\tabcolsep}r{\tabcolsep}){2-5} \cmidrule(l{\tabcolsep}r{\tabcolsep}){6-8}
            Grasp Force Controller                                           & $E$                                                       & $m$                                          & $\mu$                              & $\tau$                             & Overload [\unit{\newton}]          & Work [\unit{\milli\joule}]         & Impulse [\unit{\newton\second}]    \\
            \midrule
            \multicolumn{8}{@{}l}{\textit{Open-Loop Baseline Grasping Approaches (Non-Adaptive)}}                                                                                                                                                                                                                                                                                \\
            \, Na\"ive constant force ($3$~N)                                & ---                                                       & ---                                          & ---                                & ---                                & 2.01 {\tiny $\pm$0.41}             & 3.32 {\tiny $\pm$2.04}             & 2.15 {\tiny $\pm$1.28}             \\
            \, Class prior                                                   & 1.07 {\tiny $\pm$0.25}                                    & 1.50 {\tiny $\pm$2.13}                       & 1.21 {\tiny $\pm$0.41}             & 0.77 {\tiny $\pm$0.32}             & 1.63 {\tiny $\pm$1.42}             & 3.63 {\tiny $\pm$5.76}             & 1.95 {\tiny $\pm$2.58}             \\
            \midrule
            \multicolumn{8}{@{}l}{\textbf{\textit{Material-Aware Grasping Approaches (Adaptive), with Ablations}}}                                                                                                                                                                                                                                                               \\
            \, Ours                                                          & \textbf{0.56} {\tiny $\pm$0.23}                           & \textbf{0.49} {\tiny $\pm$0.24}              & \textbf{0.56} {\tiny $\pm$0.25}    & \textbf{0.44} {\tiny $\pm$0.17}    & \textbf{0.39} {\tiny $\pm$0.26}    & \textbf{2.25} {\tiny $\pm$5.37}    & \textbf{0.66} {\tiny $\pm$0.90}    \\
            \,\, \rotatebox[origin=c]{180}{$\Lsh$} no prior                  & \underline{0.65} {\tiny $\pm$0.81}                        & 0.62 {\tiny $\pm$0.97}                       & \underline{0.60} {\tiny $\pm$0.39}             & 0.58 {\tiny $\pm$0.20}             & 0.44 {\tiny $\pm$0.27}             & \underline{2.65} {\tiny $\pm$1.97} & \underline{0.73} {\tiny $\pm$0.62} \\
            \,\, \rotatebox[origin=c]{180}{$\Lsh$} wrong (adversarial) prior & 0.69 {\tiny $\pm$0.45}                                    & \underline{0.56} {\tiny $\pm$1.04}           & 0.63 {\tiny $\pm$0.31} & \underline{0.55} {\tiny $\pm$0.19} & \underline{0.42} {\tiny $\pm$0.33} & 2.99 {\tiny $\pm$6.76}             & 1.44 {\tiny $\pm$1.12}             \\
            \bottomrule
        \end{tabularx}
    \end{threeparttable}
    \vspace{-0.1in}
\end{table}

\begin{figure}[h!]
    \centering
    \includegraphics[width=\linewidth]{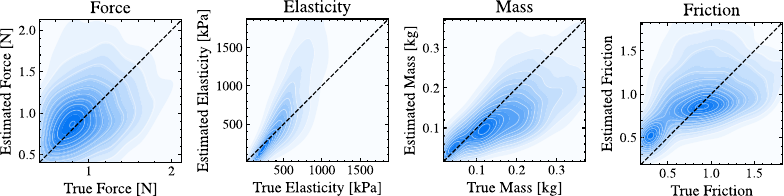}
    \caption{\textbf{Heatmap of Force and Material Estimation Errors for Real Object Grasps.} The estimated force and material properties are compared against the ground truth measurements for a set of real-world grasps of fruits and vegetables. The heatmap visualizes the distribution of estimation errors of our approach. Note that the heatmap color is indicative of the range of material properties observed in the test set, and the extent of the boundary represents the range of estimation errors using a kernel density estimate. In general, we see our approach achieves low estimation errors but may struggle at the extremes of the material property range, far from the average of the class prior.}\label{supp:fig:real-force-error-heatmaps}
    \vspace{-0.15in}
\end{figure}

\paragraph{Results in Real-World Experiments.} In our real-world experiments, our visuotactile model demonstrated a significant improvement in grasping performance compared to baseline methods. The model was able to successfully grasp a variety of test objects, applying appropriate forces that minimized the risk of damage to the objects. In \cref{supp:fig:real-force-error-heatmaps}, we show the distribution of estimation errors over the range of material properties seen in the real-world test set. While our approach performs well on average, as one would expect, we see degraded performance when the material properties are significantly different from the class prior. This matches what we observed in our ablation studies above, where we removed or adversarially adjusted the class prior.

\section{Robot Hardware and Operating System}

For both data collection and real-world validation of material-aware grasping, we utilize a 6 DoF Universal Robots UR5 arm. The arm is equipped with a Weiss Robotics WSG50-110 parallel-jaw gripper, which has a maximum jaw opening of 110~\unit{\milli\meter} and spatial resolution of 0.1~\unit{\milli\meter}. Each jaw is fitted with a 2~\unit{\kilogram} bending-beam load cell to measure the axial force applied by each gripper jaw to the object being grasped. The load cells utilize a Wheatstone bridge powered by a clean external power supply. The input voltage is regulated by LT3042 voltage regulators to ensure consistent performance, and the output voltage is amplified by INA333 voltage amplifiers. We read the differential voltage between input and output at 50~\unit{\kilo\hertz} using MCC128 DAQ hats on a Raspberry Pi 5B. At the end of each load cell, we mount a GelSight Mini tactile sensor, which provides free-running high-resolution 8~MP tactile images at a rate of 25~\unit{\Hz}.

For real-world validation of our material-aware grasping approach, we implement a ROS node that runs our learned visuotactile model for force control at 250~\unit{\Hz}. The node subscribes to the tactile image and load cell data streams, processes the data to extract the necessary features, and outputs the target grip force to the gripper control loop. We use a simple PID controller to servo the gripper towards the target force. Since GelSight Mini operates at a lower frame rate, we interpolate the tactile images to match the control frequency. The ROS node is implemented in Python and C++, and utilizes PyTorch for running the learned model inference. A desktop computer is used to read the tactile sensor data and run the control node, and it is PTP synchronized with the Raspberry Pi 5B that records and publishes the load cell measurements. We ensure that the entire pipeline from sensor data acquisition to control output operates within the required real-time constraints for effective grasping.

\end{document}